%% file: main.tex
\documentclass{article}
\usepackage[pdftex]{graphicx} 
\usepackage{subfigure} 
\graphicspath{{figures/}} 
\usepackage{natbib}
\usepackage{amsmath,amssymb}
\usepackage{graphicx}
\usepackage{algorithm} 
\usepackage{algorithmic}
\usepackage{hyperref}
\usepackage{balance}

\usepackage{bibunits}
\defaultbibliography{main}
\defaultbibliographystyle{icml2014}

\usepackage[accepted]{icml2014} 

\include{macros}

\icmltitlerunning{Stochastic Backpropagation in DLGMs}

\begin{document} 

\twocolumn[
\icmltitle{Stochastic Backpropagation and Approximate Inference \\ in Deep Generative Models}
\vspace{-3mm}
\centerline{\textbf{Danilo J. Rezende, Shakir Mohamed, Daan 
Wierstra}}
\centerline{\texttt{\{danilor, shakir, daanw\}@google.com}}
\centerline{Google DeepMind, London}

\icmlkeywords{non-linear Gaussian belief network, free energy, 
generative model, latent Gaussian models, auto-encoder}
\vskip 0.2in
]

%

\begin{abstract} 
We marry ideas from deep neural networks and approximate Bayesian inference to derive a generalised class of deep, directed generative models, endowed with a new algorithm for scalable inference and learning. 
Our algorithm introduces a recognition model to represent an approximate posterior distribution and uses this for optimisation of a variational lower bound.
We develop stochastic backpropagation -- rules for gradient backpropagation through stochastic variables -- and
 derive an algorithm that allows for joint optimisation of the parameters of both the generative and recognition models.
We demonstrate on several real-world data sets that by using stochastic backpropagation and variational inference, we obtain models that are able to
generate realistic samples of data, allow for accurate imputations of missing data, and provide a useful tool for high-dimensional data visualisation.
\end{abstract} 
\vspace{-8mm}

\begin{bibunit}[icml2013]
\input{introduction}
\input{model}
\input{stochasticBP}
\input{learning}
\input{simulations}

\input{relatedWork}
\input{conclusion}

\small{
\textit{Appendices can be found with the online version of the paper.} \url{http://arxiv.org/abs/1401.4082} }

\small{
\textbf{Acknowledgements.} 
We are grateful for feedback from the reviewers as well as Peter Dayan, Antti Honkela, Neil Lawrence and Yoshua Bengio. }

\clearpage
\small
\balance
\putbib
\end{bibunit}

\clearpage
\normalsize
\begin{bibunit}[icml2013]

\twocolumn[
\icmltitle{Appendices: \\ Stochastic Backpropagation and Approximate Inference \\ in Deep Generative Models}
\vspace{-3mm}
\centerline{\textbf{Danilo J. Rezende, Shakir Mohamed, Daan 
Wierstra}}
\centerline{\texttt{\{danilor, shakir, daanw\}@google.com}}
\centerline{Google DeepMind, London}

\icmlkeywords{non-linear Gaussian belief network, free energy, 
generative model, latent Gaussian models, auto-encoder}
\vskip 0.2in
]

\appendix

\input{appendix_estimators}
\input{appendix_VB}
\input{appendix_results}

\putbib
\end{bibunit}
\end{document}

%% file: macros.tex
\usepackage{amsmath, amsthm}
\usepackage{amsfonts}
\usepackage{amssymb}

\newcommand{\E}{\mathbb{E}} 
\newcommand{\FE}{\mathcal{F}} 

\DeclareMathOperator{\Tr}{Tr} 

  \renewenvironment{thebibliography}[1]{%
    \begin{oldthebibliography}{#1}%
      \setlength{\parskip}{0.7ex}%
      \setlength{\itemsep}{0ex}%
}{\end{oldthebibliography}}
  
\newcommand{\model}{DLGM\hspace{1mm}} 
\newcommand{\hid}{\ensuremath{\mathbf{h}}} 
\newcommand{\vis}{\ensuremath{\mathbf{v}}} 
\newcommand{\N}{\mathcal{N}} 

\definecolor{darkblue}{rgb}{0,0.05,0.35}
\definecolor{darkgreen}{rgb}{0,0.6,0}

\newcommand{\expect}[2]{\mathbb{E}_{#1}\left[#2\right]}

\newcommand{\myvec}[1]{\mbox{$\mathbf{#1}$}}
\newcommand{\myvecsym}[1]{\mbox{$\boldsymbol{#1}$}}

\newcommand{\vzero}{\mbox{$\myvecsym{0}$}}

\newcommand{\vepsilon}{\mbox{$\myvecsym{\epsilon}$}}

\newcommand{\vmu}{\mbox{$\myvecsym{\mu}$}}

\newcommand{\vtheta}{\mbox{$\myvecsym{\theta}$}}

\newcommand{\vxi}{\mbox{$\myvecsym{\xi}$}}

\newcommand{\vb}{\mbox{$\myvec{b}$}}

\newcommand{\vd}{\mbox{$\myvec{d}$}}

\newcommand{\vg}{\mbox{$\myvec{g}$}}
\newcommand{\vh}{\mbox{$\myvec{h}$}}

\newcommand{\vu}{\mbox{$\myvec{u}$}}
\newcommand{\vv}{\mbox{$\myvec{v}$}}

\newcommand{\vz}{\mbox{$\myvec{z}$}}

\newcommand{\vA}{\mbox{$\myvec{A}$}}

\newcommand{\vC}{\mbox{$\myvec{C}$}}
\newcommand{\vD}{\mbox{$\myvec{D}$}}

\newcommand{\vG}{\mbox{$\myvec{G}$}}
\newcommand{\vH}{\mbox{$\myvec{H}$}}
\newcommand{\vI}{\mbox{$\myvec{I}$}}

\newcommand{\vR}{\mbox{$\myvec{R}$}}
\newcommand{\vS}{\mbox{$\myvec{S}$}}

\newcommand{\vV}{\mbox{$\myvec{V}$}}
\newcommand{\vW}{\mbox{$\myvec{W}$}}

\newtheorem{theorem}{Theorem}[section]

\newtheorem{proposition}[theorem]{Proposition}

%% file: introduction.tex
\section{Introduction}
\label{sec:introduction}
There is an immense effort in machine learning and statistics to develop 
accurate and scalable probabilistic models of data. Such models are called 
upon whenever we are faced with tasks requiring probabilistic reasoning, 
such as prediction, missing data imputation and uncertainty estimation; or 
in simulation-based analyses, common in many scientific fields such as genetics, 
robotics and control that require generating a large number of 
independent samples from the model.

Recent efforts to develop generative models have focused on directed models, 
since samples are easily obtained by ancestral sampling from the generative 
process. Directed models such as belief networks and similar latent variable 
models \citep{dayan1995, frey1996, saul1996mean, book:bartholomew, 
uria2013, gregor2013} can be easily sampled from, but in most cases, efficient 
inference algorithms have remained elusive.
These efforts, combined with the demand for accurate probabilistic inferences 
and fast simulation, lead us to seek generative models that are
i) \textit{deep}, since hierarchical architectures allow us to capture complex 
structure in the data,
ii)  allow for \textit{fast sampling} of fantasy data from the inferred model, and
iii) are computationally \textit{tractable and scalable} to high-dimensional 
data. 

We meet these desiderata by introducing a class of deep, directed generative models with Gaussian latent variables at each layer. To allow for efficient and tractable inference, we 
use introduce an approximate representation of the posterior over the latent variables using a recognition model that acts as a stochastic encoder of the data.
For the generative model, we derive the objective function for optimisation using variational principles; for the recognition model, we specify its structure and regularisation by exploiting recent advances in deep learning. 
Using this construction, we can train the entire model by a modified form of gradient 
backpropagation that allows for optimisation of the parameters of  both the 
generative and recognition models jointly. 

We build upon the large body of prior work (in section 
\ref{sec:relatedWork}) and make the following contributions:
\begin{itemize}
\vspace{-4mm}
\setlength{\itemsep}{1pt}\setlength{\parskip}{0pt} \setlength{\parsep}{0pt}
\item We combine ideas from deep neural networks and probabilistic latent 
variable modelling to derive a general class of deep, non-linear latent 
Gaussian models (section \ref{sec:model}).
\item We present a new approach for scalable variational inference that 
allows for joint optimisation of both variational and model parameters by 
exploiting the properties of latent Gaussian distributions and gradient 
backpropagation (sections \ref{sec:stochBP} and \ref{sec:learning}).
\item We provide a comprehensive and systematic evaluation of the model 
demonstrating its applicability to problems in simulation, visualisation, 
prediction and missing data imputation (section \ref{sec:Results}).
\end{itemize}

%% file: model.tex
\vspace{-3mm}
\section{Deep Latent Gaussian Models}
\label{sec:model}
\begin{figure}[tb]
	\centering
	\subfigure[]{
		\includegraphics[width=0.22\columnwidth]{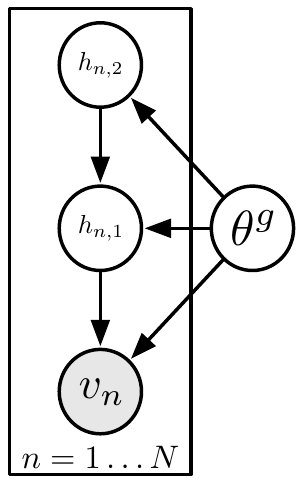}
		\label{fig:graphModel}}
	\subfigure[]{
		\includegraphics[width=0.7\columnwidth]{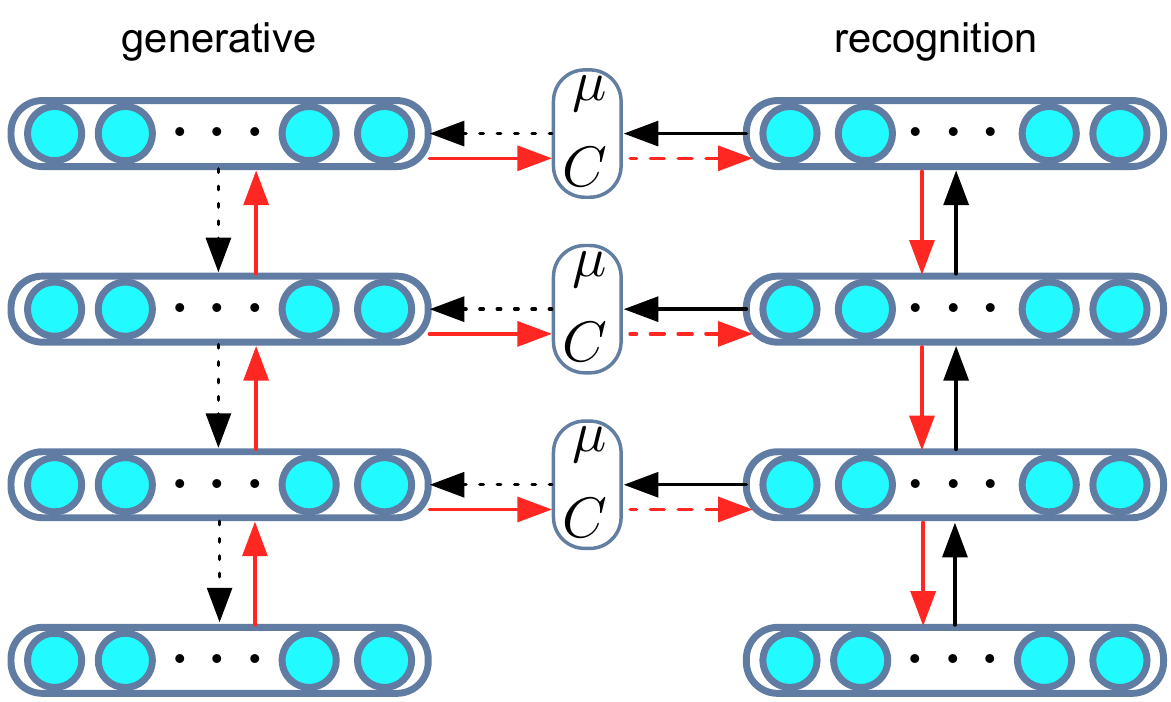} 
		\label{fig:introCompGraph}}
	\vskip -0.08in 
	\caption{(a) Graphical model for DLGMs 
		\eqref{eq:LogLikeStandard}. 
		(b) The corresponding computational graph. 
		Black arrows indicate the forward pass of 
		sampling 
		from the recognition and generative models: Solid lines indicate propagation of deterministic activations,  dotted lines 
		indicate propagation of samples. 
		Red arrows indicate the 
		backward pass for gradient computation: Solid lines indicate paths where deterministic backpropagation is used, 
		dashed arrows indicate stochastic backpropagation.}
	\label{figure:introModel}
	\vspace{-5mm}
\end{figure}

Deep latent Gaussian models (DLGMs) are a general class of deep directed graphical models that consist of Gaussian latent variables at each layer of a processing hierarchy.
The model consists of $L$ layers of latent variables. To generate a sample from the model, we begin at the top-most layer ($L$) by drawing from a Gaussian distribution. 
The activation $\vh_{l}$ at any lower layer is formed by a non-linear 
transformation of the layer above $\vh_{l+1}$, perturbed by Gaussian 
noise. 
We descend through the hierarchy and generate observations $\vv$ by 
sampling from the observation likelihood using the activation of the lowest layer 
$\vh_1$. This process is 
described graphically in figure \ref{fig:graphModel}.

This generative process is described as follows:
\vspace{-2mm}
\begin{align}
\vxi_l & \sim \N(\vxi_l | \vzero, \vI), \quad l = 1, \ldots, L \\
\hid_{L} &= \vG_{L} \vxi_{L}, \label{eq:xL} \\
\hid_{l} &= T_l(\hid_{l+1}) + \vG_{l} \vxi_{l}, \quad l =1 \ldots L-1 
\label{eq:HiddenGen} \\
\vis & \sim  \pi(\vis | T_0(\hid_{1})),
\label{eq:Visible}
\end{align}
where $\vxi_l$ are mutually independent Gaussian variables. The 
transformations $T_l$ represent multi-layer perceptrons (MLPs) and 
$\vG_l$ are matrices. At the visible layer, the data is generated from any 
appropriate distribution $\pi(\vis | 
\cdot)$ whose parameters are specified by a transformation of the first 
latent layer. 
Throughout the paper we refer 
to the set of parameters in  this generative model  by $\vtheta^g$, i.e. the parameters of the maps $T_l$ and the matrices $G_l$. This 
construction allows us to make use of as many deterministic and stochastic 
layers as needed.
We adopt a 
weak Gaussian prior over $\vtheta^g$, $p(\vtheta^g) = \N(\vtheta| \vzero, 
\kappa\vI)$.

The joint probability distribution of this model can be expressed in two equivalent ways:
\begin{align}
& p(\vis,\! \hid)\! =\! p(\!\vis| \hid_1, \!\vtheta^g ) p(\hid_L | \vtheta^g) 
p(\!\vtheta^g\!)
\!\!\prod_{l=1}^{L 
- 1}\!\! p_l( \hid_l | \hid_{l\!+\!1}, \!\vtheta^g )\label{eq:LogLikeStandard} 
\\
& p(\vis,\!\vxi)\! = p(\vis| \hid_1( \vxi_{1 \ldots L} ), \vtheta^g ) 
p(\vtheta^g) 
\prod_{l=1}^{L} 
 \N(\vxi| \vzero, \vI). \label{eq:LogLike}
\end{align}
The conditional distributions $p( \hid_l | \hid_{l+1} )$ are implicitly 
defined by equation \eqref{eq:HiddenGen} and are Gaussian distributions 
with mean $\vmu_l = T_l(\hid_{l+1})$ and covariance $\vS_l = \vG_l 
\vG_l^\top$. Equation \eqref{eq:LogLike} makes explicit that this 
generative model works by applying a complex non-linear transformation to 
a spherical Gaussian distribution $p(\vxi) = \prod_{l=1}^{L} 
\N(\vxi_l| \vzero, \vI)$ such that the transformed 
distribution tries to match the empirical distribution. A graphical model 
corresponding to equation \eqref{eq:LogLikeStandard} is shown in figure
\ref{figure:introModel}(a). 

This specification for deep latent Gaussian models (DLGMs) generalises a number 
of well known models. 
When we have only one layer of latent variables and use a linear 
mapping $T(\cdot)$, we recover \textit{factor analysis} \citep{book:bartholomew} -- more general 
mappings allow for a non-linear factor analysis \citep{lappalainen2000bayesian}. When the mappings 
are of the form $T_l(\hid) = \vA_l f(\hid) + \vb_l$, for simple element-wise 
non-linearities $f$ such as the probit function or the rectified linearity, we 
recover the \textit{non-linear Gaussian belief network} \citep{frey1999a}. We describe the relationship to other existing models in 
section \ref{sec:relatedWork}. Given this specification, our key task is to 
develop a method for tractable inference. A number of 
approaches are known and widely used, and include: mean-field variational EM \citep{beal2003}; the wake-sleep algorithm 
\citep{dayan2000}; and stochastic variational methods and related control-variate 
estimators \citep{wilson1984variance, Williams1992, hoffman2012stochastic}. We 
also follow a stochastic variational approach, but shall develop an alternative to 
these existing 
inference algorithms that overcomes many of their limitations and that is both scalable and efficient.

%% file: stochasticBP.tex
\vspace{-3mm}
\section{Stochastic Backpropagation}
\label{sec:stochBP}
Gradient descent methods in latent variable models typically require 
computations of the form $\nabla_\theta \expect{q_\theta}{f(\vxi)}$, where 
the expectation is taken with respect to a distribution $q_\theta(\cdot)$ with 
parameters
$\vtheta$, and $f$ is 
a loss function that we assume to be integrable and smooth. This quantity 
is difficult to compute directly since i)  the expectation is unknown for most problems, and ii) there is an indirect dependency on the parameters of $q$ over which the expectation is taken. 

We now develop the key identities that are used to allow for efficient inference by exploiting specific properties of the problem of computing gradients through random variables. We refer to this computational strategy as \textit{stochastic backpropagation}.
\vspace{-2mm}
\subsection{Gaussian Backpropagation (GBP)}
When the distribution $q$ is a
$K$-dimensional Gaussian $\N(\vxi | \vmu, \vC)$  
the required gradients can be computed 
using the Gaussian gradient identities:
\vspace{-1mm}
\begin{align}
\nabla_{\mu_i} \E_{ \tiny{\N(\vmu,\vC)} } \left[ f(\vxi) \right ] &= \E_{ 
\tiny{\N(\vmu,\vC)} } \left[
\nabla_{\xi_i} f(\vxi) \right], \label{eq:bonnet} \\
\nabla_{C_{ij}} \E_{ \tiny{\N(\vmu,\vC) } }\left [ f(\vxi) \right ] &= \tfrac{1}{2}  \E_{ 
\tiny{\N(\vmu,\vC)} } \left [ 
\nabla^2_{\xi_i,\xi_j} f(\vxi) \right], \label{eq:price}
\end{align}
which are due to the theorems by \citet{bonnet1964} and  \citet{price1958}, 
respectively. These equations are true in expectation 
 for any integrable and smooth function $f(\vxi)$. Equation 
\eqref{eq:bonnet} is a direct consequence of the location-scale 
transformation for the Gaussian (discussed in section 
\ref{sect:stochBP_genRules}). Equation \eqref{eq:price} can be derived by 
successive application of the product rule for integrals; we provide the proofs 
for these identities in appendix \ref{app:gaussDerivProofs}. 

Equations \eqref{eq:bonnet} and \eqref{eq:price} are especially interesting 
since they allow for unbiased gradient estimates by using a 
small number of samples from $q$. 
Assume that both the mean $\vmu$ and covariance matrix $\vC$ depend on 
a parameter vector $\vtheta$. We are now able to write a general rule for 
Gaussian gradient computation by combining equations \eqref{eq:bonnet} 
and \eqref{eq:price} and using the 
chain rule:
\vspace{-1mm}
\begin{equation}
\!\!\!\nabla_{\!\tiny{\vtheta}}\E_{\! \tiny{\N\!\left(\vmu, \vC\right)} }\! \left [ f(\vxi) \right ] \!=\! 
\E_{\! \tiny{\N\!\left(\vmu, \vC\right)} } \!\!\left [\! \mathbf{g}\!^\top\! \frac{\partial 
\vmu}{\partial \vtheta}\! +\! \tfrac{1}{2}\! \Tr\! \left(\!\mathbf{H} 
\frac{\partial \vC}{\partial \vtheta} \!\right) \!\right ] \label{eq:GBP}
\end{equation}
where $\vg$ and $\vH$ are the gradient and the Hessian of the function 
$f(\vxi)$, respectively. Equation \eqref{eq:GBP} can be interpreted as a modified 
backpropagation rule for Gaussian distributions that takes into account 
the gradients through the mean $\vmu$ \textit{and} covariance $\vC$. This  
reduces to the 
standard backpropagation rule when $\vC$ is constant. Unfortunately this 
rule requires knowledge of the Hessian 
matrix of $f(\vxi)$, which has an algorithmic complexity $O(K^3)$. For inference in DLGMs, we later introduce an unbiased though higher 
variance estimator that requires only quadratic complexity.
\vspace{-3mm}
\subsection{Generalised Backpropagation Rules}
\label{sect:stochBP_genRules}
We describe two approaches to derive general backpropagation rules for non-Gaussian $q$-distributions.

\textbf{Using the product rule for integrals.}
For many exponential family distributions, it is possible to find a function 
$B(\vxi; \vtheta)$ to ensure that
\vspace{-2mm}
\begin{align}
\nabla_{\tiny{\vtheta}} \E_{\tiny{p(\vxi | \vtheta )}} [f(\vxi)] \!=\!\! =  -\E_{\tiny{p (\vxi | 
\vtheta )}} [\nabla_{\tiny{\vxi}} [B (\vxi; \vtheta) f (\vxi)]].  \nonumber
\end{align}
That is, we express the gradient with respect to the parameters of $q$ as an 
expectation of gradients with respect to the random variables themselves.
This approach can be used to derive rules for many 
distributions such as the Gaussian, inverse Gamma and log-Normal. 
We discuss this in more detail in appendix
\ref{app:otherStochBPrule}. 

\textbf{Using suitable co-ordinate transformations.} \\
We can also derive stochastic backpropagation rules for any distribution 
that can be written as a smooth, invertible  transformation of a standard base 
distribution. For example, any Gaussian distribution $\N(\vmu, \vC)$ can be 
obtained as a transformation of a spherical Gaussian $\vepsilon \sim 
\N(\vzero,\vI)$, using the transformation $y = \vmu + \vR \vepsilon$ and 
$\vC = \vR \vR\!^\top\!\!$. The gradient 
of the expectation with respect to $\vR$ is then:
\vspace{-2mm}
\begin{align}
\nabla_{\tiny{\vR}} \E_{\!\N\!\left(\tiny{\vmu,\vC}\right) } \left[ 
f(\vxi) \right ]  &=\nabla_{\tiny{\vR}} 
\E_{\tiny{\N(0,I)} } \left [ f(\small{\vmu + \vR \vepsilon}) 
\right ] \nonumber \\ 
 &= \E_{ \tiny{\N(0,I)} } \left [
\vepsilon  \vg^\top \right],  \label{eq:gradR}
\end{align}
where $\mathbf{g}$ is the gradient of $f$ evaluated at $\vmu + \vR \vepsilon$ and provides a lower-cost alternative to Price's theorem \eqref{eq:price}.
Such transformations are well known for many distributions, especially those 
with a self-similarity property or location-scale formulation, such as the 
Gaussian, Student's $t$-distribution, stable distributions, and generalised extreme value distributions.

\textbf{Stochastic backpropagation in other contexts.}
The Gaussian gradient identities described above do not appear to be widely used. These identities have been recognised by \citet{opper2009} for variational inference in Gaussian process regression, and following this work, by \citet{gravesNIPS2011} for parameter learning in large neural networks. 
Concurrently with this paper, \citet{kingma2013auto} present an alternative discussion of
stochastic backpropagation. Our approaches were developed simultaneously and provide complementary perspectives on the use and derivation of stochastic backpropagation rules. 

%% file: learning.tex
\vspace{-3mm}
\section{Scalable Inference in DLGMs}
\label{sec:learning}
We use the matrix $\vV$ to refer to the full data set of size 
$N\times D$ with observations $\vv_n = [v_{n1}, \ldots, v_{nD}]^\top$. 

\vspace{-2mm}
\subsection{Free Energy Objective}
To perform inference in DLGMs we must integrate out the effect of 
any latent variables -- this requires us to compute the integrated or marginal 
likelihood. 
In general, this will be an intractable integration and instead we optimise a 
lower bound on the marginal likelihood. We introduce an 
approximate posterior distribution $q(\cdot)$ and apply Jensen's inequality 
following the variational principle \citep{beal2003} to obtain:
\vspace{-3mm}
\begin{align}
\mathcal{L}(\vV) &= -\log p(\vV) = -\log \int p(\vV |\vxi,\vtheta^g) 
p(\vxi,\vtheta^g)  d\vxi \nonumber \\
& = -\log \int  \frac{ q(\vxi)}{q(\vxi) } p(\vV |\vxi,\vtheta^g) 
p(\vxi,\vtheta^g)
 d\vxi  \label{eq:FE} \\ 
{\leq} \FE(\vV) & \!=\!D_{KL}[q(\vxi) \| p(\vxi)]\! -\! \expect{q}{\log p(\vV 
|\vxi,\!\vtheta^g)p(\vtheta^g)} .\nonumber
\end{align}
This objective consists of two terms: the first 
is the KL-divergence between the variational distribution and the prior 
distribution (which acts a regulariser), and the second is a reconstruction 
error. 

We specify the approximate posterior as  a distribution $q(\vxi | \vv)$ that is \textit{conditioned} on the observed data. This distribution can be specified as any directed acyclic graph where each node of the graph is a Gaussian conditioned, through linear or non-linear transformations, on its parents. The joint distribution in this case is non-Gaussian, but stochastic backpropagation can still be applied.  

For simplicity, we use a $q(\vxi | \vv)$ that is a Gaussian distribution that factorises across 
the $L$ layers (but not necessarily within a layer):
\vspace{-2mm}
\begin{equation}
q(\vxi | \vV, \vtheta^r) = \prod_{n=1}^N \prod_{l=1}^L  \N\left( 
\vxi_{n,l} | 
\vmu_{l}(\vis_n), \vC_l(\vis_n)  \right),
\label{eq:Q}
\end{equation}
where the mean $\vmu_l(\cdot)$ and covariance $\vC_l(\cdot)$ are generic maps represented by 
deep neural networks. Parameters of the $q$-distribution are denoted by the vector $\vtheta^r$.

For a Gaussian prior and a Gaussian recognition model, the KL term in \eqref{eq:FE} can be computed analytically and the free energy becomes:
\vspace{-1mm}
\begin{align}
D_{\!KL}&[\N(\vmu,\!\vC)\| \N(\vzero, \!\vI)] \!=\! \tfrac{1}{2}\! 
\left[\textrm{Tr}(\!\vC\!)\!-\!\log|\vC| \!+\! \vmu\!^\top\!\!\vmu \!-\!D \right], 
\nonumber \\
\FE(\vV) &= -\sum_n
\E_{q} \left[ \log p(\vis_n| \hid(\vxi_n) ) \right]  
 + \tfrac{1}{2 \kappa} \Vert \vtheta^g\Vert ^2 \nonumber \\
&+\! \frac{1}{2}\! \sum_{n,l}\! \left[ \Vert \vmu_{n,l} \Vert ^2 \!+ \!
\Tr(\vC _{n,l}) \! -\! \log \vert \vC_{n,l} \vert \!-\!1 \right]\!,\! 
\label{eq:FEsimp}
\end{align}
where $\Tr(\vC)$ and $\vert \vC \vert$ indicate the trace and the 
determinant of the covariance matrix $\vC$, respectively.

The specification of an approximate posterior distribution that is conditioned on the observed data is the first component of an efficient variational inference algorithm. We shall refer to the distribution $q(\vxi| \vv)$ \eqref{eq:Q} as a \textit{recognition model}, whose design is independent of the generative model. A recognition model allows us introduce a form of amortised inference \citep{gershman2014} for variational methods in which we share statistical strength by allowing for generalisation across the posterior estimates for all latent variables using a model. The implication of this generalisation ability is: faster convergence during training; and faster inference at test time since we only require a single pass through the recognition model, rather than needing to perform any iterative computations (such as in a generalised E-step). 

To allow for the best possible inference, the specification of the recognition model must be flexible enough to provide an accurate approximation of the posterior distribution -- motivating the use of deep neural networks. 
We regularise the recognition model by introducing additional noise, specifically, bit-flip or drop-out noise at the input layer and small additional Gaussian noise to samples from the recognition model. 
We use rectified linear activation functions as non-linearities for any deterministic layers of the neural network. We found that such regularisation is essential and without it the recognition model is unable to provide accurate inferences for unseen data points. 

%
\vspace{-2mm}
\subsection{Gradients of the Free Energy}
To optimise \eqref{eq:FEsimp}, we use Monte Carlo methods for any expectations and use stochastic 
gradient descent for optimisation. For optimisation, we  require efficient estimators of the gradients 
of all 
terms in equation \eqref{eq:FEsimp} with respect to the parameters $\vtheta^g$ and $\vtheta^r$
of the generative  and the recognition models, respectively.

The gradients with respect to the $j$th generative parameter $\theta_j^g$  can 
be computed using:
\vspace{-2mm}
\begin{align}
\nabla_{ \theta_j^g } \FE (\vV)  &= -  \E_{q} \left[ \nabla_{\theta_j^g} \log 
p(\vV| \hid ) 
\right] +\tfrac{1}{\kappa}  \theta_j^g. \label{eq:dFEdThetaG}
\end{align}
An unbiased estimator of $\nabla_{ \theta_j^g } \FE (\vV)$ is obtained by 
approximating equation \eqref{eq:dFEdThetaG} with a small number of samples 
(or 
even a single sample)
from the recognition model $q$.

To obtain gradients with respect to the recognition parameters $\vtheta^r$, 
we use the rules for Gaussian backpropagation developed in section 
\ref{sec:stochBP}. To address the complexity of the Hessian in the general 
rule 
\eqref{eq:GBP}, we use the co-ordinate transformation for the Gaussian to 
write the gradient with respect to the factor matrix $\vR$ instead of the 
covariance $\vC$ (recalling $\vC = \vR \vR^\top$) derived in equation 
\eqref{eq:gradR}, where derivatives are computed for the function $f(\vxi) = \log p(\vis| \hid(\vxi))$.

The gradients of $\FE(\vis)$ in equation \eqref{eq:FEsimp} with 
respect to the variational mean $\vmu_l (\vis)$ and the factors 
$\vR_l(\vis)$ 
are:
\vspace{-2mm}
\begin{align}
\nabla_{\mu_l} \FE(\vis) 
&= -  \E_{q} \left[ \nabla_{\vxi_l} \log p(\vis| \hid(\xi) ) 
\right] + \vmu_l, \label{eq:dFEdMu} \\
\nabla_{R_{l,i,j}} \FE(\vis)
&= - \tfrac{1}{2}\E_{q} \left[ \epsilon_{l,j}
\nabla_{\xi_{l,i}} \log p(\vis| \hid(\vxi) ) 
\right] \nonumber\\
&+ \tfrac{1}{2}\nabla_{R_{l,i,j}}  \left[ \Tr \vC _{n,l}  - \log \vert \vC_{n,l} 
\vert \right], \label{eq:dFEdR}
\end{align}
where the gradients $\nabla_{R_{l,i,j}}  \left[ \Tr \vC _{n,l}  - \log \vert 
\vC_{n,l} 
\vert \right]$ are computed by backpropagation.
Unbiased estimators of the gradients \eqref{eq:dFEdMu} and 
\eqref{eq:dFEdR} are obtained jointly by 
sampling from the recognition model $\vxi \sim q(\vxi | \vis)$ (bottom-up 
pass) and updating the values of the generative model layers using 
equation \eqref{eq:HiddenGen} (top-down 
pass). 

Finally the gradients $\nabla_{ \theta_j^r } \FE (\vis)$ obtained from 
equations \eqref{eq:dFEdMu} and \eqref{eq:dFEdR} are:
\vspace{-2mm}
\begin{align}
\!\!\!\nabla_{\small{\vtheta^r}} \FE (\vis) \!=\! 
\! \nabla_{\small{\vmu}} \FE (\vis)\!^\top\! \frac{\partial \vmu}{\partial 
\vtheta^r}\! +\! \! \Tr \left(\!\nabla_{\small{\vR}} \FE(\vis)\frac{\partial 
\vR}{\partial 
\vtheta^r} \!\right) \!. \label{eq:dFEdThetaR}
\end{align}
The gradients \eqref{eq:dFEdThetaG} -- 
\eqref{eq:dFEdThetaR} are now used to descend the free-energy surface with respect to  
both the 
generative and recognition parameters in a single optimisation step. 
Figure \ref{fig:introCompGraph} shows the flow of computation in 
DLGMs. Our algorithm proceeds by first performing a forward pass (black 
arrows), consisting of a bottom-up (recognition) phase and a top-down 
(generation) phase, which updates the hidden activations of the recognition 
model and parameters of any Gaussian distributions, and then a backward 
pass 
(red arrows) in which gradients are computed using the appropriate 
backpropagation rule for deterministic and stochastic layers.
We take a descent step using:
\vspace{-1mm}
\begin{equation} 
\Delta \theta^{g,r} = -\Gamma^{g,r} {\nabla_{\theta^{g,r} } 
\FE(\vV)}, \label{eq:SGD}
\vspace{-1mm}
\end{equation}
where $\Gamma^{g,r}$ is a diagonal pre-conditioning matrix computed using the RMSprop 
heuristic\footnote{Described by G. Hinton, 
	`RMSprop: Divide the gradient by a running average of its recent 
		magnitude', in \textit{Neural networks for machine learning}, 
	Coursera lecture 6e, 2012.}. The learning 
procedure is summarised in algorithm 
\ref{alg:train}.
\vspace{1.5mm}
\begin{algorithm}[tbp]
  \caption{Learning in {\model}\!s}
  \label{alg:train}
\begin{algorithmic}
\WHILE{ hasNotConverged() }
\STATE $\vV \gets \text{getMiniBatch}()$
\STATE $\vxi_n \sim q(\xi_n | \vis_n)$ (bottom-up pass) eq. \eqref{eq:Q}
\STATE $\hid \leftarrow \hid(\xi)$ (top-down pass) eq. \eqref{eq:HiddenGen}
\STATE updateGradients() eqs \eqref{eq:dFEdThetaG} -- 
\eqref{eq:dFEdThetaR}
\STATE $\vtheta^{g,r} \gets \vtheta^{g,r} + \Delta \vtheta^{g,r}$
\ENDWHILE
\end{algorithmic}
\end{algorithm}

\vspace{-3mm}
\subsection{Gaussian Covariance Parameterisation}
\label{sec:learning_covParam}

There are a number of approaches for parameterising the covariance matrix 
of the recognition model $q(\vxi)$. Maintaining a full covariance matrix 
$\vC$ in 
equation 
\eqref{eq:FEsimp} would 
entail an algorithmic complexity of $O(K^3)$ for training and sampling per layer, 
where $K$ is the number of latent variables per layer.

The simplest approach is to use a diagonal covariance matrix $\vC = 
\textrm{diag}(\vd)$, where $\vd$ is a $K$-dimensional vector. This 
approach is appealing since it allows for linear-time computation and 
sampling, but only allows for axis-aligned posterior distributions.

We can improve upon the diagonal approximation by 
parameterising the covarinace as a rank-1 matrix with a diagonal correction. 
Using a vectors $\vu$ and $\vd$, with $\vD = \textrm{diag}(\vd)$, we parameterise the precision 
 $\vC^{-1}$ as:
\vspace{-2mm}
\begin{equation}
\vC^{-1} = \vD + \vu \vu^\top. \label{eq:Cinv}
\vspace{-2mm}
\end{equation}
This representation allows for arbitrary rotations of the Gaussian distribution 
along one principal direction with relatively few additional parameters 
\citep{magdon-ismail2010a}. By application of the matrix inversion lemma (Woodbury identity), we obtain 
the covariance matrix in terms of $\vd$ and $\vu$ as:
\vspace{-2mm}
\begin{align}
& \vC  = \vD^{-1} - \eta \vD^{-1} \vu \vu^\top \vD^{-1}, \label{eq:C}  & 
\quad \eta = \tfrac{1}{\vu^\top \vD^{-1} \vu + 1}, \nonumber \\
& \log \vert \vC \vert  = \log \eta - \log \vert \vD \vert. 
\vspace{-2mm}
\end{align}
This allows both the trace $\Tr(\vC)$  and $\log \vert \vC \vert$ needed 
in the computation of the Gaussian KL, as well as their gradients, to be computed in $O(K)$ time per layer.

The factorisation $\vC = \vR \vR^\top$, with $\vR$ a matrix of the same size as 
$\vC$ and can be computed directly in terms of $\vd$ and $\vu$. One solution for $\vR$ is:
\begin{align}
\vR = \vD^{ -\frac{1}{2} } - \left[ \frac{1 - \sqrt{\eta}}{\vu^\top \vD^{-1} 
\vu} \right] \vD^{-1} \vu \vu^\top \vD^{ -\frac{1}{2} }.
\label{eq:R}
\end{align}
The product of $\vR$ with an arbitrary vector can be 
computed in $O(K)$ without computing $\vR$ explicitly. This also
allows 
us to sample efficiently from this Gaussian, since any Gaussian 
random variable $\vxi$ with mean $\vmu$ and covariance matrix $\vC = 
\vR \vR^\top$ can be written as
$\vxi = \vmu + \vR \vepsilon, \label{eq:XiR}$
where $\vepsilon$ is a standard Gaussian variate. 

Since this covariance parametrisation has linear cost in the number of latent 
variables, we can also use it to 
parameterise the variational distribution of all layers jointly, instead of the 
factorised assumption in \eqref{eq:Q}. 

\subsection{Algorithm Complexity}
\label{sec:learning_algo}

The computational complexity of producing a sample from the 
generative model is $O(L\bar{K}^2)$,  where 
$\bar{K}$ is the average number of latent variables per layer and $L$ is the 
number of layers (counting both deterministic and stochastic layers). The computational 
complexity per training sample during training is also $O(L\bar{K}^2)$ --  the same as that of matching  auto-encoder.

%% file: simulations.tex
\vspace{-3mm}
\section{Results}
\label{sec:Results}
Generative models have a number of applications in simulation, prediction, 
data visualisation, missing data imputation and other forms of probabilistic 
reasoning. We describe the testing methodology we use and present results 
on a number of these tasks. 
\vspace{-2mm}

\subsection{Analysing the Approximate Posterior}
\begin{figure*}[tbp]
	\centering
	\subfigure[Diagonal covariance]{
		\includegraphics[height = 2cm]{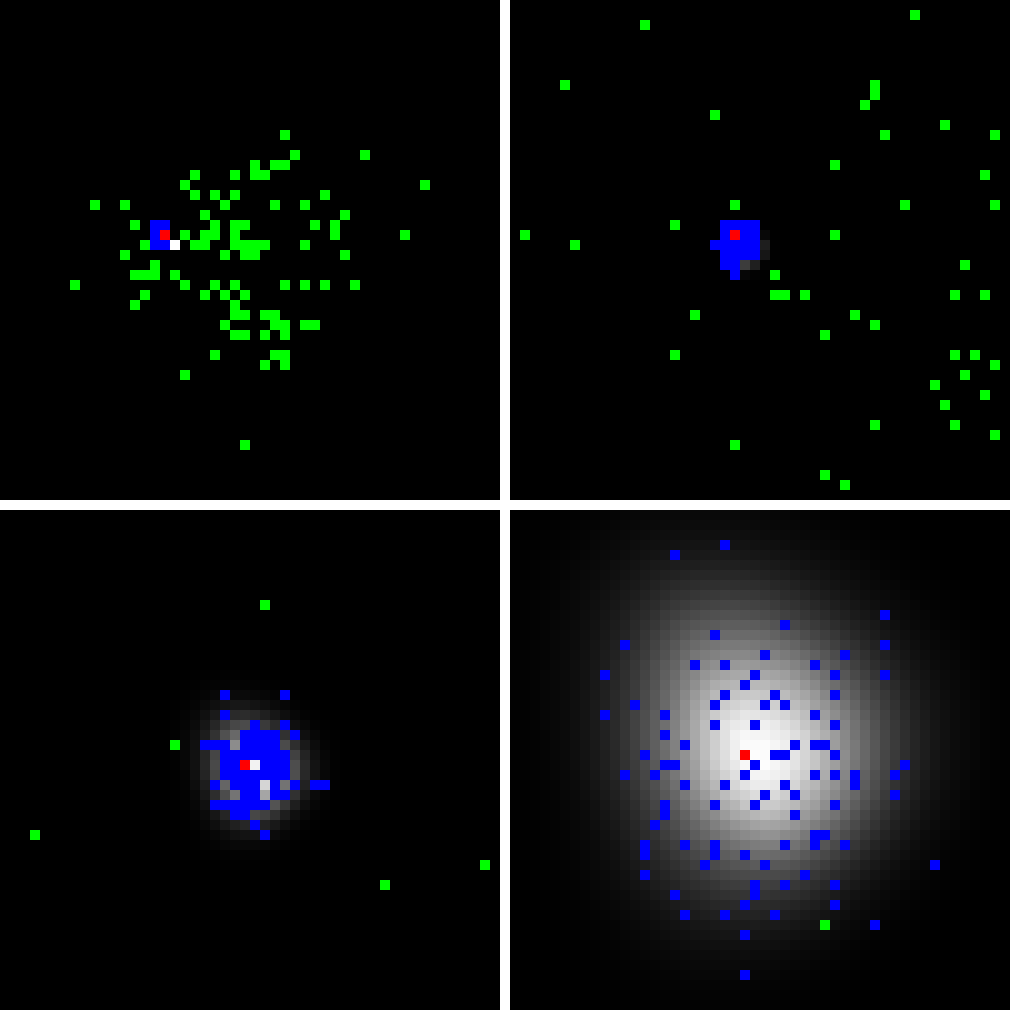}
		\includegraphics[height = 2cm]{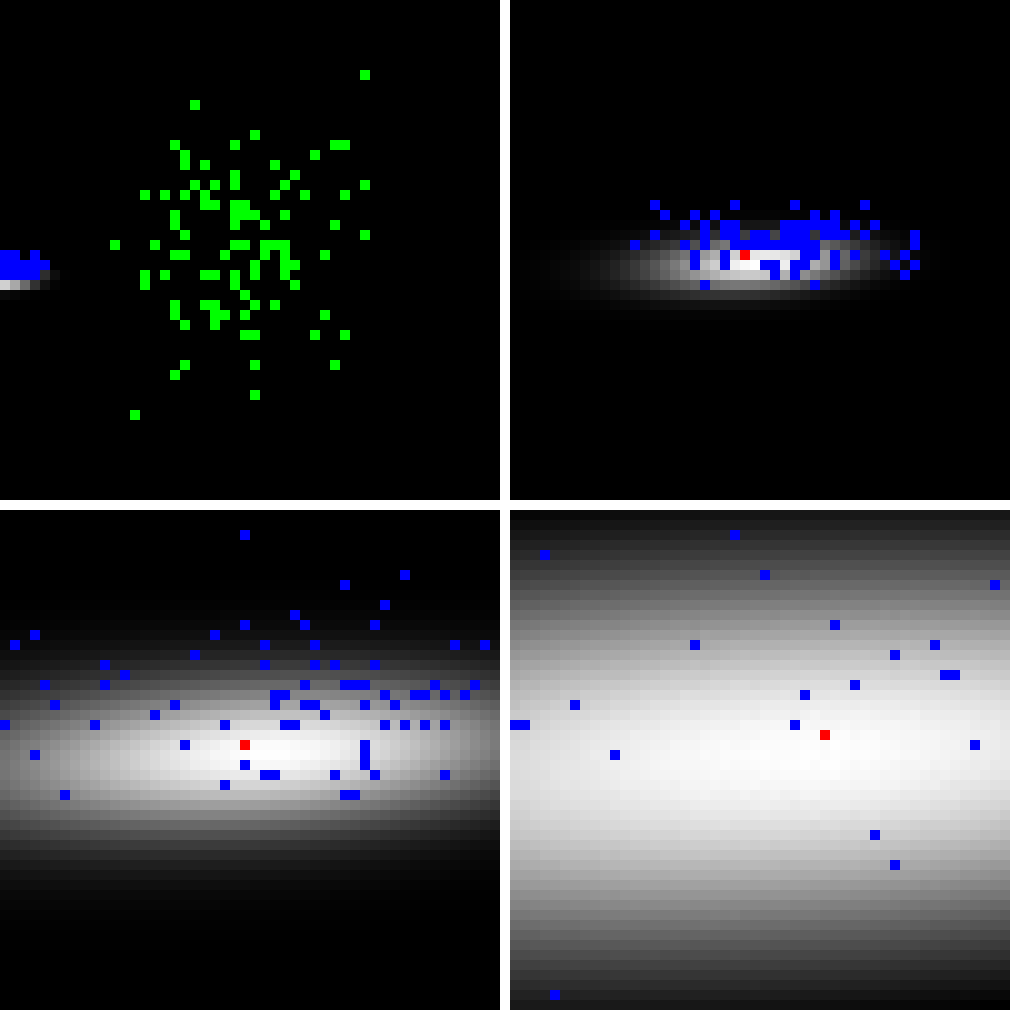}
		\includegraphics[height = 2cm]{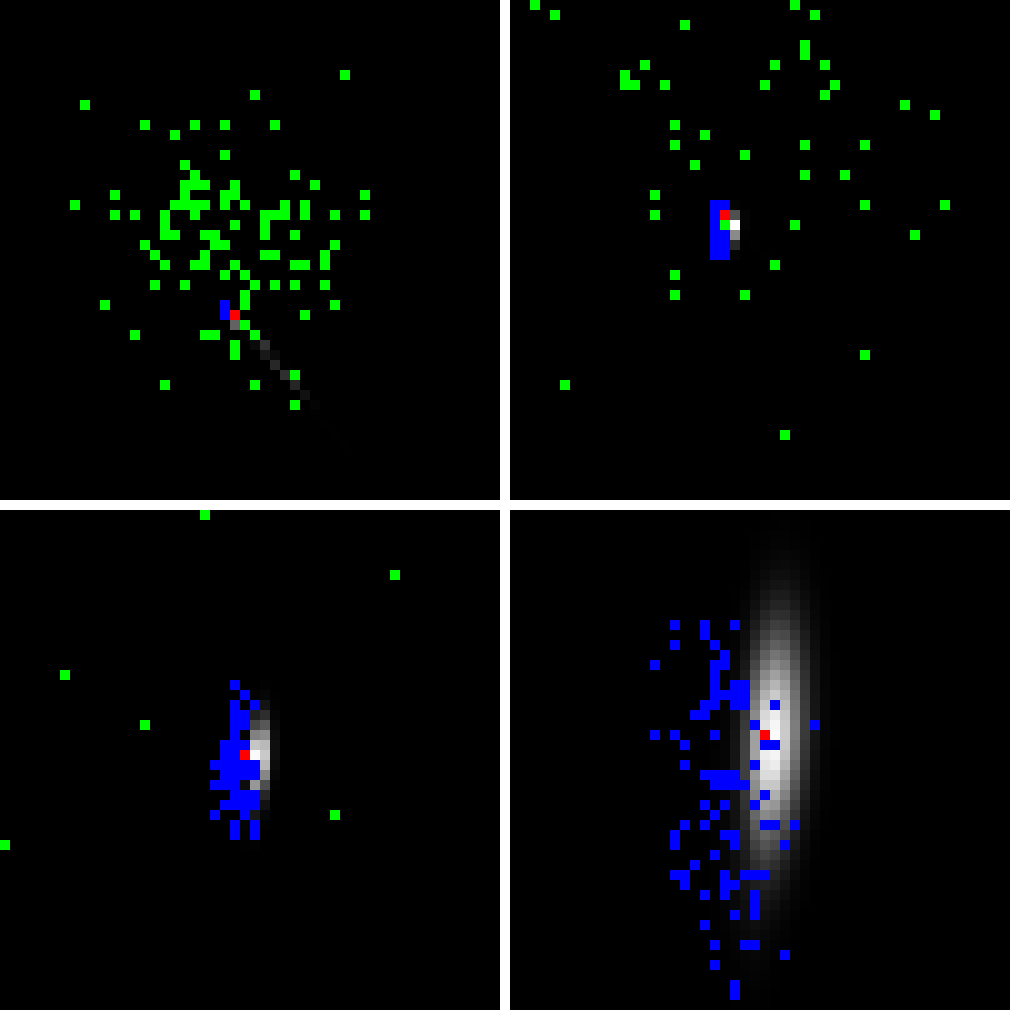}}
	\hspace{0.25cm}
	\subfigure[Low-rank covariance]{
		\includegraphics[height = 2cm]{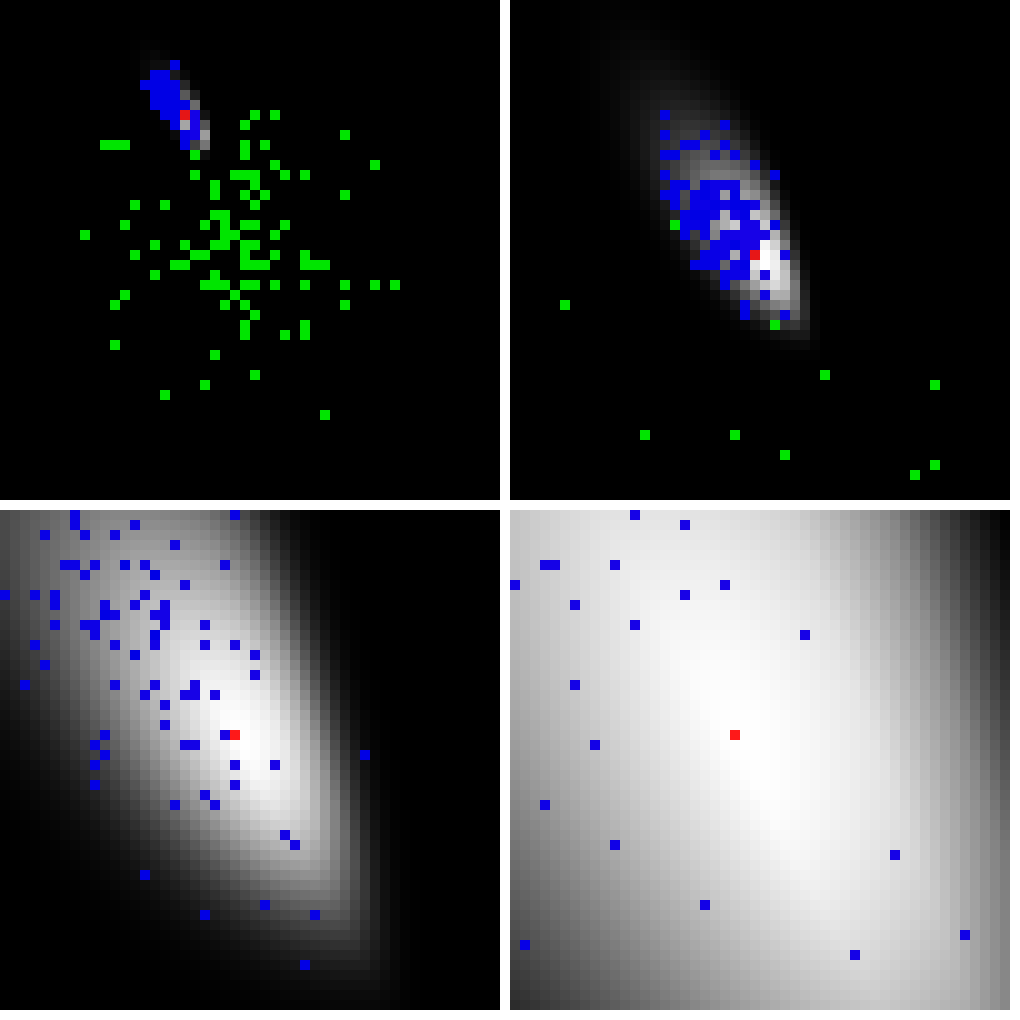}
		\includegraphics[height = 2cm]{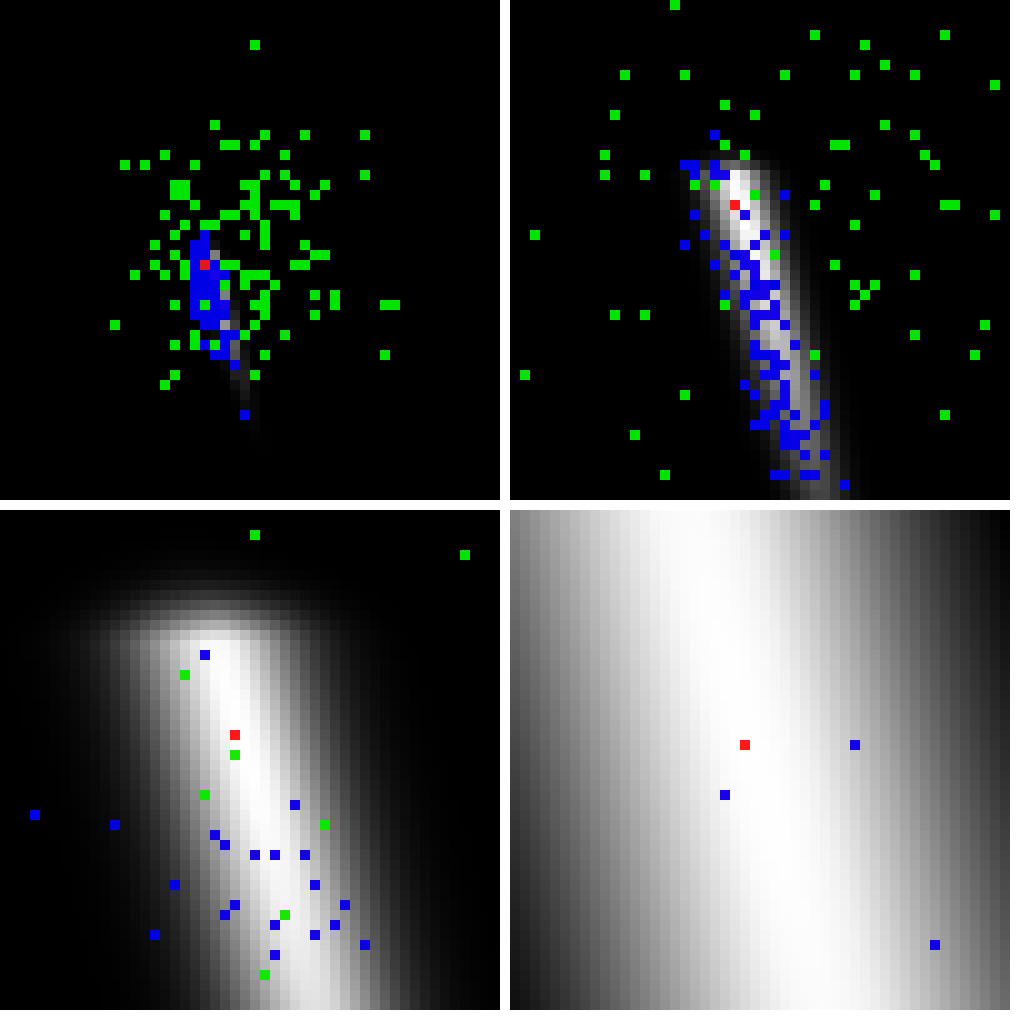}
		\includegraphics[height = 2cm]{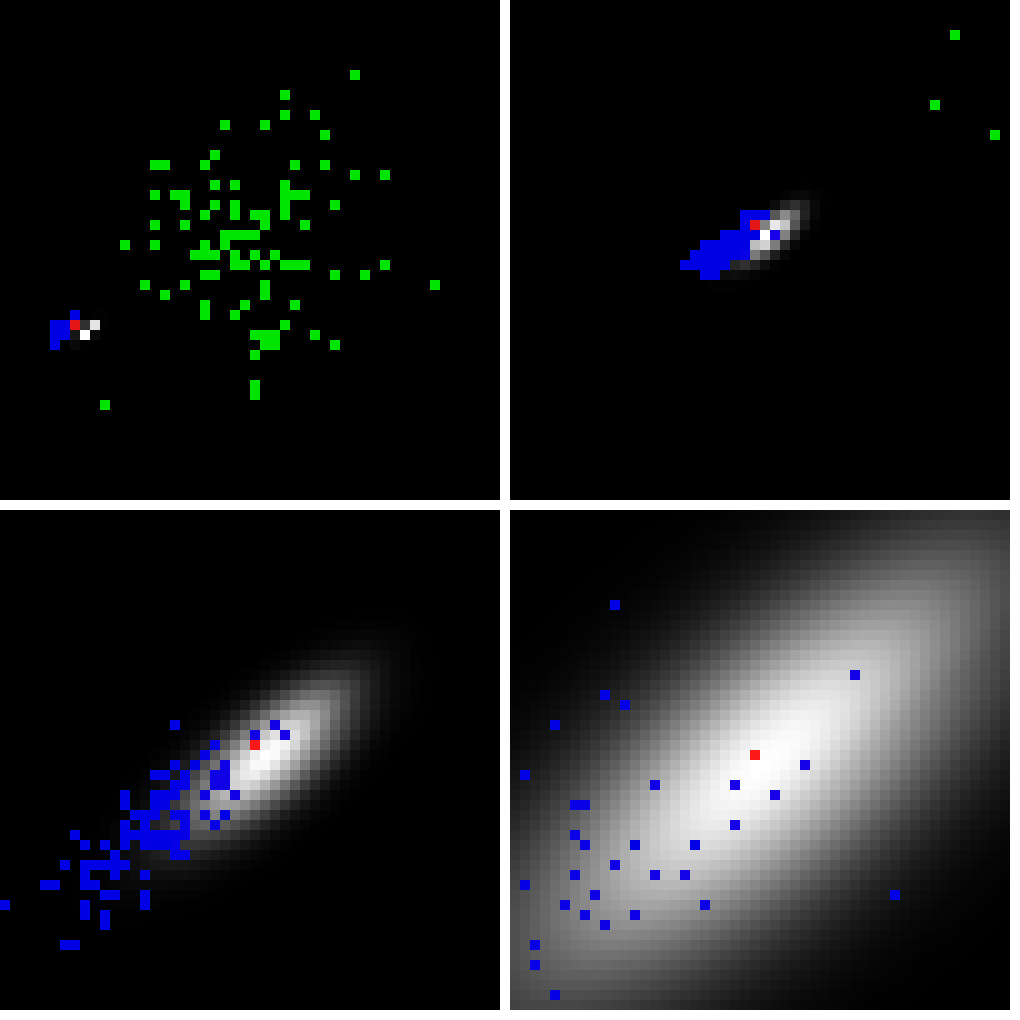}}
	\hspace{0.25cm}
	\subfigure[Performance]{
		\includegraphics[height = 2cm]{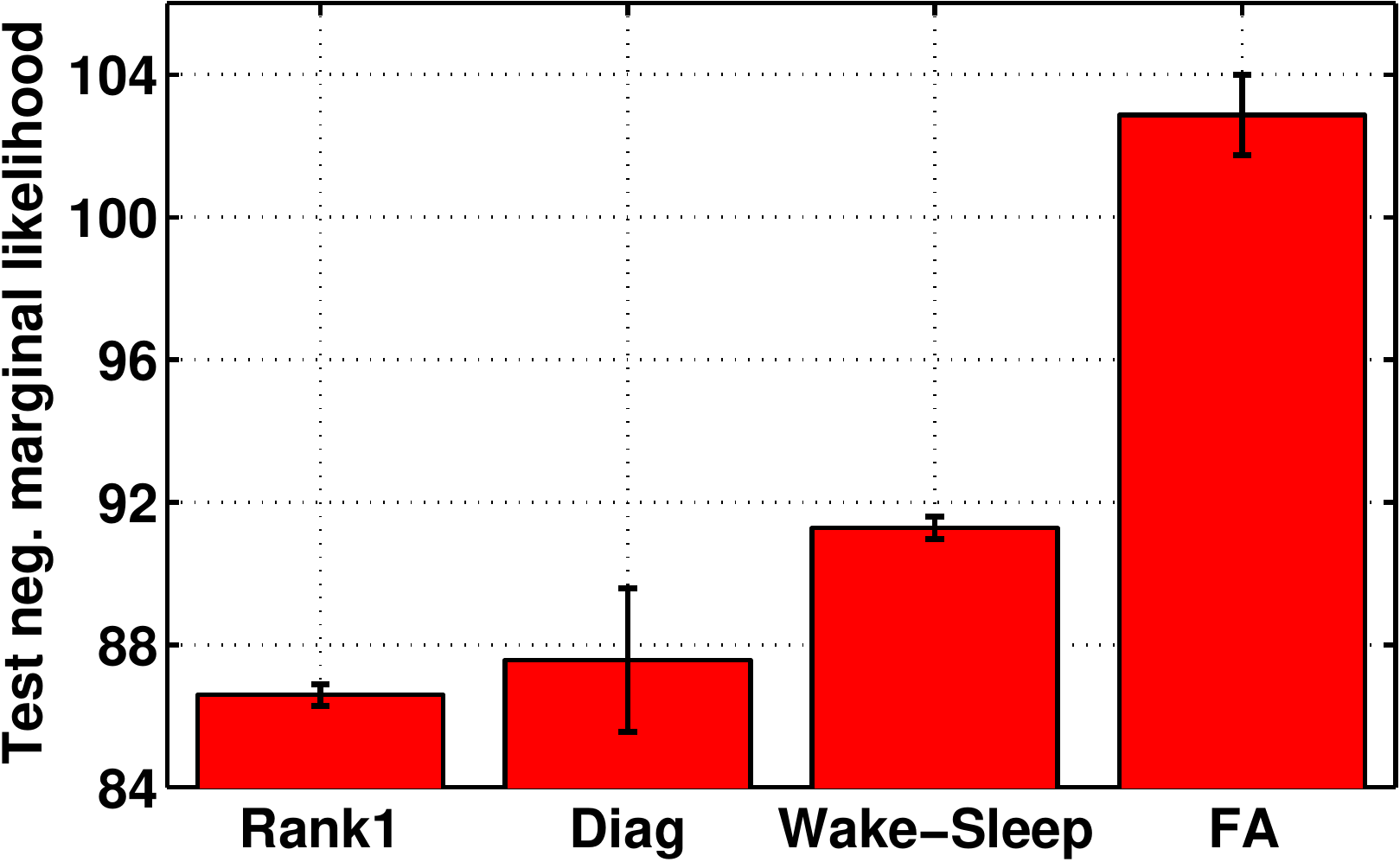}}
	\vskip-0.14in
	\caption{(a, b) Analysis of the true  vs. approximate posterior for
		MNIST. 
		Within each image we show four views of the same posterior, 
		zooming in on the region centred on the MAP (red) estimate. 
		(c) Comparison of test log likelihoods. 
		}
	\label{figure:MNIST_POSTERIOR}
	\vspace{-3mm}
\end{figure*} 


We use sampling to evaluate the true posterior distribution for a number of 
MNIST digits using the binarised data set from \citet{larochelle2011}.
We visualise the posterior distribution for a 
model with two Gaussian latent variables in figure 
\ref{figure:MNIST_POSTERIOR}. The true posterior distribution is shown by 
the grey regions and was computed by importance sampling with a large 
number of particles aligned in a grid between -5 and 5. In figure 
\ref{figure:MNIST_POSTERIOR}(a) we see that these posterior distributions  
are elliptical or spherical in shape and thus, it is reasonable to assume that 
they can be well approximated by a Gaussian. Samples from the 
prior (green) are spread widely over the space and very few 
samples fall in the region of significant posterior mass, explaining the 
inefficiency of estimation methods that rely on samples from the prior. 
Samples from the recognition model (blue) are concentrated on 
the posterior mass, indicating that the recognition model has learnt the 
correct posterior statistics, which should lead to efficient learning. 
\begin{table}[tbp]
\vspace{-3mm}
\centering
\scriptsize
\caption{Comparison of negative log-probabilities on the test set for the binarised MNIST data. }
\label{tab:mnist}
\begin{tabular}{lc}
\hline
\textbf{Model} & $-\ln p(\mathbf{v})$ \\
\hline \hline
Factor Analysis & $106.00$ \\
NLGBN \citep{frey1999a} & $95.80$ \\
Wake-Sleep \citep{dayan2000} & 91.3\\
DLGM diagonal covariance & $87.30$ \\
DLGM rank-one covariance & $86.60$ \\
\hline
\multicolumn{2}{c}{\tiny\textit{Results below from \citet{uria2013}}} \\
MoBernoullis K=10 & 168.95 \\
MoBernoullis K=500 & 137.64 \\
RBM (500 h, 25 CD steps) approx. & 86.34\\
DBN 2hl approx. & 84.55\\
NADE 1hl (fixed order) & 88.86\\
NADE 1hl (fixed order, RLU, minibatch) & 88.33\\
EoNADE 1hl (2 orderings) & 90.69 \\
EoNADE 1hl (128 orderings) & 87.71 \\
EoNADE 2hl (2 orderings) & 87.96 \\
EoNADE 2hl (128 orderings) & 85.10 \\
\hline
\end{tabular}
\vspace{-7mm}
\end{table}
\newline \newline In figure \ref{figure:MNIST_POSTERIOR}(a) we see that samples from the 
recognition model are aligned to the axis and do not capture the posterior 
correlation. The correlation is captured using the structured covariance model
in
figure \ref{figure:MNIST_POSTERIOR}(b).
Not all posteriors are Gaussian in shape, but the recognition places mass in 
the best location possible to provide a reasonable approximation. As a benchmark for comparison, the performance in terms of test log-likelihood 
is shown in figure \ref{figure:MNIST_POSTERIOR}(c), using the same 
architecture, for factor 
analysis (FA), the wake-sleep algorithm, 
and our approach using both the diagonal and structured covariance approaches. For this experiment, the generative model consists of 100 latent variables feeding into a deterministic layer of 300 nodes, which then feeds to the observation likelihood. We use the same structure for the recognition model.
\subsection{Simulation and Prediction}
\begin{figure*}[tbp]
	\centering
	\subfigure[Left: Training 
	data. Middle: Sampled pixel probabilities. Right: Model samples]{
		\frame{
			\includegraphics[height=3.8cm]{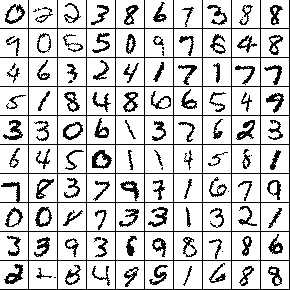}}
		\frame{
			\includegraphics[height=3.8cm]{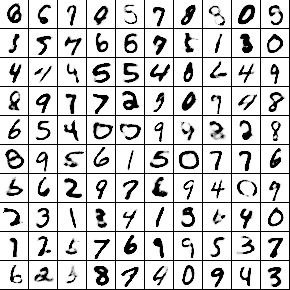}}
		\frame{
			\includegraphics[height=3.8cm]{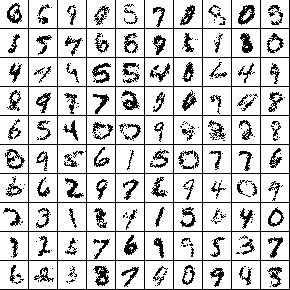}}
		\label{figure:MNIST}
	}
	\hspace{0.5cm}
	\subfigure[2D embedding.]{
		\includegraphics[height = 3.8cm]{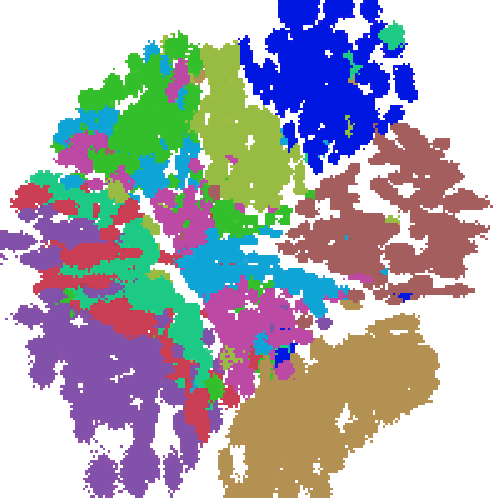}
		\label{fig:mnist_embedding}}
	\vspace{-4mm}
	\caption{Performance on the MNIST dataset. For the visualisation, each colour corresponds to one of the digit classes.}
	\vspace{-2mm}
\end{figure*} 
\begin{figure*}[tbp]
	\centering
	\subfigure[NORB]{
		\includegraphics[height = 2.6cm]{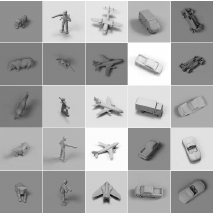}
		\includegraphics[height = 2.6cm]{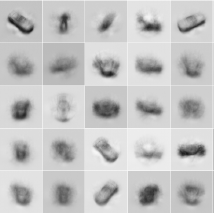}
		\label{figure:NORB}
	}
	\subfigure[CIFAR]{
		\includegraphics[height = 2.6cm]{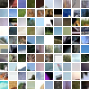}
		\includegraphics[height = 2.6cm]{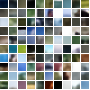}
		\label{figure:CIFAR10}
	}
	\subfigure[Frey]{
		\includegraphics[height=2.6cm]{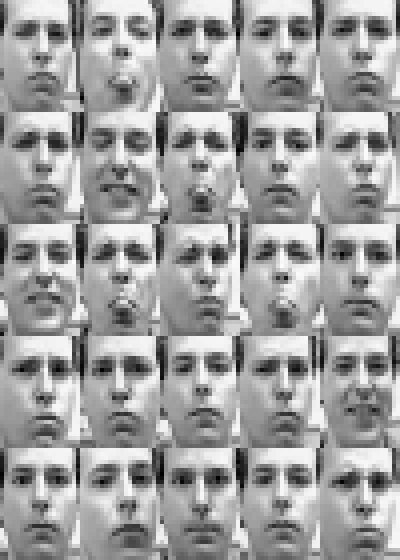}
		\includegraphics[height=2.6cm]{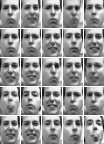}
		\label{fig:freyFaces}
	}
	\vspace{-3mm}
	\caption{Sampled generated from DLGMs for three data sets: (a) NORB, (b) CIFAR 10, (c) Frey faces. In all images, the left image shows samples from the 
		training data and the right side shows the generated samples.}
	\vspace{-3mm}
\end{figure*}
We evaluate the performance of a three layer latent Gaussian model on the 
MNIST data set. The model consists of two deterministic layers with 200 
hidden units and a stochastic layer of 200 latent variables. We use 
mini-batches of 200 observations and trained the model using stochastic backpropagation. 
Samples from this model are shown in figure \ref{figure:MNIST}. We also 
compare the test log-likelihood to a large number of existing approaches 
in table \ref{tab:mnist}. We used the binarised dataset as in 
\citet{uria2013} and quote the log-likelihoods in the lower part of the 
table from this work. These results show that our approach is competitive with 
some of the best models currently available. The generated digits also 
match the true data well and visually appear as good as some of the best 
visualisations from these competing approaches. 

We also analysed the performance of our model on three high-dimensional 
real image data sets. The NORB object recognition data set consists of 
$24,300$ images that are of size $96 \times 96$ pixels. We use a 
model consisting of 1 deterministic layer of 400 hidden units and 
one stochastic layer of 100 latent variables. Samples produced from this 
model are shown in figure \ref{figure:NORB}. The CIFAR10 natural images 
data set consists of $50,000$ RGB images that are of size $32 
\times 32$ pixels, which we split into random $8 \times 8$ patches. We use 
the same 
model as used for the MNIST experiment and show samples from the model 
in figure \ref{figure:CIFAR10}. The Frey faces data set consists of almost 
$2,000$ images of different facial expressions
of size 
$28 \times 20$ pixels. 

\subsection{Data Visualisation}
Latent variable models are often 
used for visualisation of 
high-dimensional data sets. We project the MNIST data set to a 
2-dimensional latent space and use this 2D embedding as a visualisation 
of the data --  an embedding for MNIST is shown in 
figure \ref{fig:mnist_embedding}. The classes separate into different 
regions,  suggesting that such embeddings can be useful in understanding the structure of high-dimensional data sets.

\subsection{Missing Data Imputation and Denoising}
\begin{figure}[tb]
\centering
\frame{\includegraphics[width = 8cm]{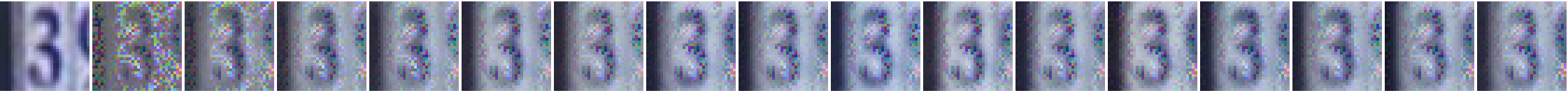}}
\frame{\includegraphics[width = 8cm]{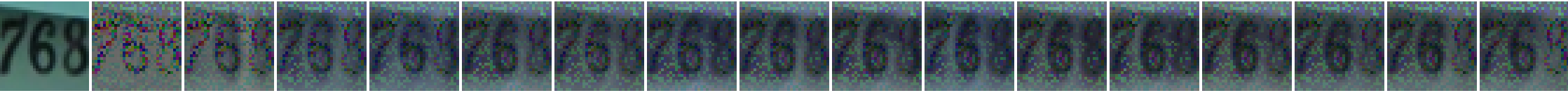}}
\frame{\includegraphics[width = 8cm]{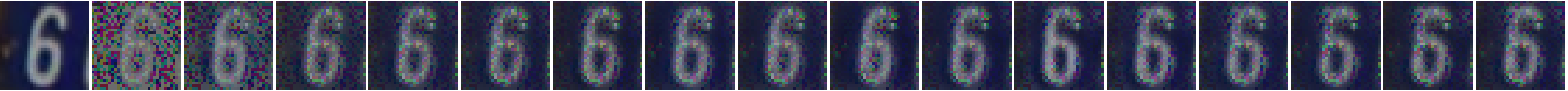}}
\vskip 4px
\frame{\includegraphics[width = 8cm]{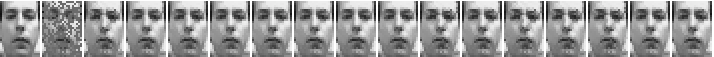}}
\vskip 4px
\frame{\includegraphics[width = 8cm]{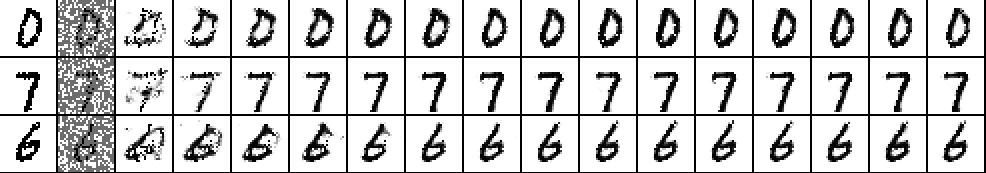}}
\vskip 4px
\frame{\includegraphics[width = 8cm]{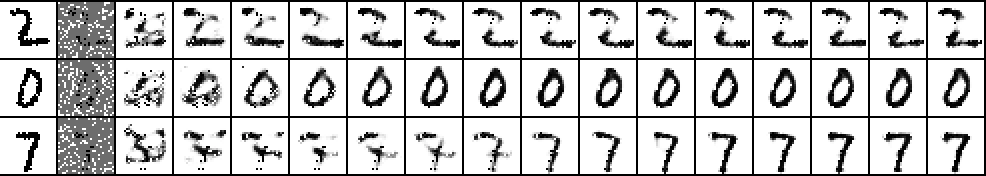}}
\vskip 4px
\frame{\includegraphics[width = 8cm]{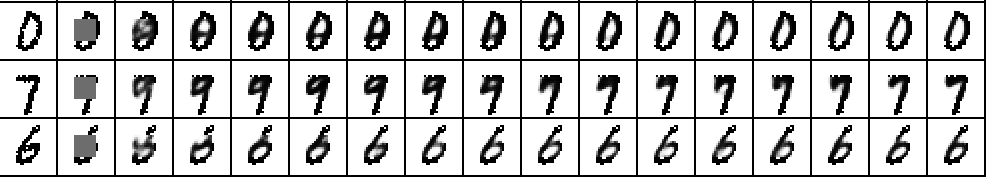}}
\caption{Imputation results:  Row 1, SVHN. Row 2, Frey faces.
	Rows 3--5, MNIST. Col. 1 shows the true data. Col. 2 shows pixel locations set as missing in grey. The remaining columns show imputations for 15 iterations. 
	}
\label{fig:mnist_completion}
\vspace{-6mm}
\end{figure}

We demonstrate the ability of the model to impute missing data using the street view house numbers (SVHN) data set \citep{SVHNdata}, which consists of $73,257$ images of size $32 \times 32$ pixels, and the Frey faces and MNIST data sets. The performance of the model is shown in figure \ref{fig:mnist_completion}. 

We test the imputation ability under two different missingness types 
\citep{little1987statistical}: Missing-at-Random (MAR), where we consider 
60\% and 80\% of the pixels to be missing randomly, and Not 
Missing-at-Random (NMAR), where we consider a square region of the image 
to be missing.
The model produces very good completions in both test cases. There is uncertainty in the identity of the image and this is
reflected in the errors in these completions as the resampling procedure is 
run (see transitions from digit 9 to 7, and digit 8 to 6 in figure 
\ref{fig:mnist_completion} ). This further demonstrates the ability of the 
model to capture the diversity of the underlying data. 
We do not integrate over the missing values, 
but use a procedure that simulates a Markov chain that we show converges 
to the true marginal distribution of missing given observed pixels. The 
imputation procedure is discussed in appendix \ref{ap:ImageCompletion}.

%% file: relatedWork.tex
\section{Discussion}
\label{sec:discussion}
\label{sec:relatedWork}
\textbf{Directed Graphical Models.} 
DLGMs form a unified family of models that includes factor analysis \citep{book:bartholomew}, non-linear factor analysis \citep{lappalainen2000bayesian}, and non-linear Gaussian belief networks \citep{frey1999a}.
Other related models include
sigmoid belief networks \citep{saul1996mean} and
deep 
auto-regressive networks \citep{gregor2013}, which use 
auto-regressive 
Bernoulli distributions at each 
layer instead of Gaussian distributions. The Gaussian process latent 
variable model and deep Gaussian processes
\citep{lawrence2005probabilistic, damianou2013} form the non-parametric analogue of 
our model and employ Gaussian process priors over the non-linear functions 
between each layer. 
The neural auto-regressive density estimator (NADE) \citep{larochelle2011, uria2013} 
uses function approximation to model conditional distributions within a 
directed acyclic graph. 
NADE is amongst the most competitive generative models currently available, 
but has several limitations,
such as the inability to allow for deep representations and difficulties in 
extending to locally-connected models (e.g., through the use of 
convolutional layers),  preventing it from scaling easily to 
high-dimensional data.

\textbf{Alternative latent Gaussian inference.} Few of the alternative approaches for inferring latent Gaussian distributions meet the desiderata for scalable inference we seek. 
The Laplace approximation has been concluded to be a poor approximation in general, in addition to being computationally expensive.
INLA is restricted to models with few hyperparameters ($<10$), whereas our interest is in 100s-1000s.
EP cannot be applied to latent variable models due to the inability to match moments of the joint distribution of latent variables and model parameters. 
Furthermore, no reliable methods exist for moment-matching with means and covariances formed by non-linear transformations -- linearisation and importance sampling are two, but are either inaccurate or very slow. Thus, the the variational approach we present remains a general-purpose and competitive approach for inference.

\textbf{Monte Carlo  variance reduction.} 
\textit{Control variate} methods are amongst the most 
general and 
effective techniques for variance reduction when Monte Carlo methods are used \citep{wilson1984variance}. One 
popular approach is the
REINFORCE algorithm \citep{Williams1992}, since it is simple to implement 
and applicable to both discrete and continuous  
models, though control variate methods are becoming increasingly popular 
for variational inference problems \citep{hoffman2012stochastic, 
	blei2012variational, ranganath2013, salimans2014}. Unfortunately, such 
estimators  have the undesirable property 
that their variance scales linearly with the number of independent random 
variables in the target function, while the variance of GBP is bounded by a 
constant: for $K$-dimensional latent variables the variance of 
REINFORCE scales as $O(K)$, whereas GBP scales as $O(1)$ (see appendix \ref{ap:var_reinforce}).

An important family of alternative estimators is based on \textit{quadrature and series 
	expansion methods} \citep{HonkelaV04, lappalainen2000bayesian}. These methods
have low-variance at the price of introducing biases in the estimation. More 
recently a combination of the series expansion and control variate 
approaches has been proposed by \citet{blei2012variational}.

A very general alternative 
is the \textit{wake-sleep algorithm} 
\citep{dayan1995}.  The wake-sleep algorithm can perform well, but it fails to optimise a single consistent objective function and there is 
thus no guarantee that optimising it leads to a decrease in the free energy (\ref{eq:FE}). 

\textbf{Relation to denoising auto-encoders.} 
Denoising auto-encoders (DAE) \citep{vincent2010} introduce a 
random corruption to the encoder network and attempt to minimize the 
expected  reconstruction error under this corruption noise with additional 
regularisation terms. In our variational approach, the 
recognition distribution $q(\vxi|\vis)$ can be interpreted as a stochastic 
encoder in the DAE setting. There is then a direct correspondence between the expression for the free energy 
(\ref{eq:FE}) and the reconstruction 
error and regularization terms used in denoising auto-encoders (c.f. 
equation (4) of \citet{bengio2013}). Thus, we can see denoising auto-encoders as a realisation of 
variational inference in latent variable models.

The key difference is that the 
form of encoding `corruption' and regularisation terms used in our model 
have been derived directly using the variational principle to provide a strict 
bound on the marginal likelihood of a known directed graphical model 
that allows for easy generation of samples. 
DAEs can also be used as generative models by simulating from a Markov chain \citep{bengio2013,bengio2013deep}.
But the behaviour of these Markov chains will be very problem specific, 
and we lack consistent tools to evaluate their convergence. 




%% file: conclusion.tex
\vspace{-3mm}
\section{Conclusion}
\label{sec:Discussion}

%
%
We have introduced a general-purpose inference method for 
models with continuous latent variables. 
Our approach introduces a recognition model, which can be seen as a stochastic encoding of the data, to allow for efficient and tractable inference. 
We derived a lower bound on the marginal likelihood for the generative model and specified the structure and regularisation of the recognition model by exploiting recent advances in deep learning. 
By developing modified rules for backpropagation through stochastic 
layers, we derived an efficient inference algorithm that allows for joint 
optimisation of all parameters.
We show on several real-world data sets that the
model generates realistic samples, provides accurate imputations of 
missing data and can be a useful tool for high-dimensional data visualisation. 

%% file: appendix_estimators.tex
\section{Additional Model Details}
\label{app:altView}
In equation \eqref{eq:LogLike} we showed an alternative form of the joint 
log likelihood that explicitly separates the 
deterministic and stochastic parts of the generative model and corroborates 
the view that the generative model works by 
applying a complex non-linear transformation to a spherical Gaussian 
distribution $\N(\vxi| \vzero, \vI)$ such that the transformed distribution 
best 
matches the empirical distribution. We provide more details on this view 
here for clarity.

From the model description in equations \eqref{eq:HiddenGen} and 
\eqref{eq:Visible}, we can interpret the variables $\hid_l$ as deterministic 
functions of the noise variables $\vxi_l$.  This can be formally introduced 
as a coordinate transformation of the probability density in equation 
\eqref{eq:LogLikeStandard}: we perform a change of coordinates $\hid_l 
\rightarrow \vxi_{l}$. The density of the transformed variables 
$\vxi_l$ can be expressed in terms of the density (\ref{eq:LogLikeStandard}) 
times the determinant of the Jacobian of the transformation $p(\vxi_l)  = 
p(\hid_l(\vxi_l)) 
|\frac{\partial \hid_l}{\partial \vxi_l}| $. Since the co-ordinate transformation is 
linear we have $|\frac{\partial \hid_l}{\partial \vxi_l}| = |\vG_l|$ and the distribution 
of $\vxi_l$ is obtained as follows:
\begin{align}
&p(\vxi_l)\!\!  = 
p(\hid_l(\vxi_l)) 
|\frac{\partial \hid_l}{\partial \vxi_l}| \nonumber \\
& p(\vxi_l)\!\! = \! p(\hid_L) \! |\vG_L| \!
\prod_{l=1}^{L - 1} \!  |\!\vG_l\!| p_l( \hid_l | \hid_{l+1}\! ) 
\! = \!\! \prod_{l=1}^{L} \! |\vG_l| |\vS_l|^{-\frac{1}{2}}  \N( \vxi_l ) 
\nonumber \\
&=  \prod_{l=1}^{L}  |\vG_l| |\vG_l \vG_l^T|^{-\frac{1}{2}} \N( \vxi_l | 
\vzero, \vI) = 
\prod_{l=1}^{L} \N\!( \vxi_l | \vzero, \vI).
\end{align}

Combining this equation with the distribution of the visible layer we obtain 
equation \eqref{eq:LogLike}. 

\subsection{Examples}

Below we provide simple, explicit examples of generative and recognition 
models.

In the case of a two-layer model the activation $\hid_1(\vxi_{1,2})$ in 
equation \eqref{eq:LogLike} can be 
explicitly written as
\begin{eqnarray}
\hid_1(\vxi_{1,2}) & = \vW_1 f ( \vG_2 \vxi_2 ) + \vG_1 \vxi_1 + 
\vb_1.
\end{eqnarray}

Similarly, a simple recognition model consists of a single 
deterministic layer and a stochastic Gaussian layer with the rank-one 
covariance structure and is constructed as:
\begin{eqnarray}
& q(\vxi_l | \vis) = \N\left(\vxi_l | \vmu; (\textrm{diag}(\vd) +  \vu \vu^\top)^{-1}\right) \\
& \vmu = \vW_\mu \vz + \vb_\mu \\
& \log \vd =  \vW_d \vz + \vb_d  ; \qquad
\vu = \vW_u \vz + \vb_u \\
&\vz = f(\vW_v \vv + \vb_v)
\end{eqnarray}
where the function $f$ is a rectified linearity (but other non-linearities 
such as tanh can be used).

\makeatletter{\renewcommand*{\@makefnmark}{}
\footnote{\textit{Proceedings of the
$\mathit{31}^{st}$ International Conference on Machine Learning},
Beijing, China, 2014. JMLR: W\&CP volume 32. 
Copyright 2014 by the author(s).} \makeatother}

\section{Proofs for the Gaussian Gradient Identities}
\label{app:gaussDerivProofs}

Here we review the derivations of Bonnet's and Price's theorems that were presented in section \ref{sec:stochBP}.

\begin{theorem}[Bonnet's theorem]
Let $f(\vxi):  R^d  \mapsto R$ be a integrable and twice differentiable 
function. The gradient of the expectation of $f(\vxi)$ under a Gaussian 
distribution $\mathcal{N}(\vxi| \vmu,\vC)$ with respect to the mean $\vmu$ 
can be expressed as the 
expectation of the gradient of $f(\vxi)$.
\begin{align}
\nabla_{\mu_i} \E_{ \N(\mu,\vC) } \left[ f(\vxi) \right ] &= \E_{ \N(\vmu,\vC) } 
\left[
\nabla_{\xi_i} f(\vxi) \right],\nonumber
\end{align}
\end{theorem}

\begin{proof}
\begin{align}
\nabla_{\mu_i} \E_{ \N(\vmu,\vC) } \left[ f(\vxi) \right ] &=  \int \nabla_{\mu_i} 
\N(\vxi| 
\vmu,\vC)  f(\vxi) d\vxi \nonumber \\
&= -\int \nabla_{\xi_i} 
\N(\vxi| 
\vmu,\vC)  f(\vxi) d\vxi \nonumber  \\
&= \left[ \int  \N(\vxi| 
\vmu,\vC) f(\vxi) d \vxi_{\neg i} \right]_{ \xi_i = -\infty }^{\xi_i =  + \infty} 
\nonumber \\
&+  \int 
\N(\vxi| 
\vmu,\vC) \nabla_{\xi_i}   f(\vxi) d\vxi \nonumber  \\
&= \E_{ \N(\vmu,\vC) } 
\left[
\nabla_{\xi_i} f(\vxi) \right],
\end{align}
where we have used the identity
\begin{align}
\nabla_{\mu_i} \N(\vxi| \vmu,\vC) &= - \nabla_{\xi_i} 
\N(\vxi| \vmu,\vC) \nonumber 
\end{align}
in moving from step 1 to 2. From step 2 to 3 we have used the product 
rule for integrals with the first term evaluating to zero.
\end{proof}

\begin{theorem}[Price's theorem]
Under the same conditions as before. The gradient of the expectation of 
$f(\vxi)$ under a Gaussian 
distribution $\mathcal{N}(\vxi| \vzero,\vC)$ with respect to the covariance 
$\vC$ can be expressed in terms of the 
expectation of the Hessian of $f(\vxi)$ as 
\begin{align}
\nabla_{C_{i,j}} \E_{ \N(\vzero,C) } \left[ f(\vxi) \right ] &=  \frac{1}{2} \E_{ 
\N(\vzero,C) } 
\left[
\nabla_{\xi_i, \xi_j} f(\vxi) \right] \nonumber
\end{align}
\end{theorem}

\begin{proof}
\begin{align}
\nabla_{C_{i,j}} \E_{ \N(\vzero,\vC) } \left[ f(\vxi) \right ] &=  \int 
\nabla_{C_{i,j}} 
\N(\vxi| \vzero,\vC)  f(\vxi) d\vxi \nonumber \\
&= \frac{1}{2} \int \nabla_{\xi_i, \xi_j}
\N(\vxi| \vzero,\vC)  f(\vxi) d\vxi \nonumber \\
&= \frac{1}{2} \int 
\N(\vxi| \vzero,\vC)  \nabla_{\xi_i, \xi_j} f(\vxi) d\vxi \nonumber \\
&=  \frac{1}{2} \E_{ 
\N(\vzero,\vC) } 
\left[
\nabla_{\xi_i, \xi_j} f(\vxi) \right].
\end{align}
In moving from steps 1 to 2, we have used the identity
\begin{align}
\nabla_{C_{i,j}} \N(\vxi| \mu,\vC) &= \frac{1}{2} \nabla_{\xi_i, \xi_j} 
\N(\vxi| \mu,\vC), \nonumber 
\end{align}
which can be verified by taking the derivatives on both sides and comparing 
the resulting expressions. From step 2 to 3 we have used the 
product 
rule for integrals twice.
\end{proof}

\section{Deriving Stochastic Back-propagation Rules}
\label{app:otherStochBPrule}
In section \ref{sec:stochBP} we described two ways in which to derive 
stochastic back-propagation rules. We show specific examples and provide 
some more discussion in this section.
\subsection{Using the Product Rule for Integrals}
We can derive rules for stochastic back-propagation for many distributions by finding a appropriate non-linear function $B(x; \theta)$ that allows us to express the gradient with respect to the parameters of the distribution as a gradient with respect to the random variable directly. The approach we described in the main text was:
\begin{align}
& \nabla_{\theta} \E_{p} [f(x)] \!=\!\! 
\int \!\nabla_{\theta} p (\!x | \theta )\! f\!(\!x\!) d x  
\!=\!\! \int \!\nabla_x p (\!x | \theta ) B (\!x\!) f\!(\!x\!) d x \nonumber \\
& =  [B (x) f (x) p ( x | \theta)]_{supp (x)} - \int p
(x | \theta) \nabla_x [B (x) f (x)]   \nonumber \\
& =  -\E_{p (x | \theta )} [\nabla_x [B (x) f (x)]]  \label{eq:app_prodRuleSBP}
\end{align}
where we have introduced the non-linear function $B(x;\theta)$ to allow for the transformation of the gradients and have applied the product rule for integrals (rule for integration by parts) to rewrite the integral in two parts in the second line, and the supp$(x)$ indicates that the term is evaluated at the boundaries of the support. To use this approach, we require that the density we are analysing be zero at the boundaries of the support to ensure that the first term in the second line is zero.

As an alternative, we can also write this differently and find an non-linear function of the form:
\begin{align}
\nabla_{\theta} \E_{p} [f(x)] \!=\!\! =  -\E_{p (x | \theta )} [B (x) \nabla_x  f (x)]. 
\end{align}

Consider general exponential family distributions of the form:
\begin{align}
p (x | \theta) = h (x) \exp (\eta (\theta)^\top \phi (x) - A(\theta))
\end{align}
where $h(x)$ is the base measure, $\theta$ is the set of mean parameters of the distribution, $\eta$ is the set of natural parameters, and $A(\theta)$ is the log-partition function. We can express the non-linear function in \eqref{eq:app_prodRuleSBP} using these quantities as:
\begin{align}
B(x) = \frac{[\nabla_{\theta} \eta (\theta) \phi (x) - \nabla_{\theta} A
(\theta)]}{[\nabla_x \log [h (x)] + \eta (\theta)^T \nabla_x \phi
(x)]}.
\end{align}
This can be derived for a number of distributions such as the Gaussian, inverse Gamma, Log-Normal, Wald (inverse Gaussian) and other distributions. We show some of these below:
\begin{table}[h]
\centering
\small
\begin{tabular}{l c c}
\hline
\textbf{Family} & $\theta$ & $B(x)$ \\
\hline \hline
Gaussian & $\left(\begin{array}{c}\mu\\\sigma^2 \end{array}\right)$ & $\left(\begin{array}{c}
  - 1\\
   \frac{(x - \mu - \sigma^{}) (x - \mu + \sigma)^{}}{2 \sigma^2 (x -
  \mu)}
\end{array}\right)$\\
Inv. Gamma & $\left(\begin{array}{c}\alpha\\\beta \end{array}\right)$ & $\left(\begin{array}{c} \frac{x^2(- \ln x - \Psi (\alpha) + \ln \beta)}{- x (\alpha + 1) + \beta}\\
  (\frac{x^2}{- x (\alpha + 1) + \beta})(- \frac{1}{x} + \frac{\alpha}{\beta})
\end{array}\right)$\\
Log-Normal & $\left(\begin{array}{c}\mu\\\sigma^2 \end{array}\right)$  & $\left(\begin{array}{c}
  - 1\\
   \frac{(\ln x - \mu - \sigma^{}) (\ln x - \mu + \sigma)^{}}{2
  \sigma^2 (\ln x - \mu)}
\end{array}\right)$ \\
\hline
\end{tabular}
\end{table}

The $B(x ;\theta)$ corresponding to the second formulation can also be derived and may be useful in certain situations, requiring the solution of a first order differential equation. This approach of searching for non-linear transformations leads us to the second approach for deriving stochastic back-propagation rules.

\subsection{Using Alternative Coordinate Transformations}
There are many distributions outside the exponential family that we would 
like to consider using. A simpler approach is to search for a co-ordinate 
transformation that allows us to separate the deterministic and stochastic 
parts of the distribution. We described the case of the Gaussian in section 
\ref{sec:stochBP}. Other distributions also have this property. As an 
example, consider the Levy distribution (which is a special case of the 
inverse Gamma considered above). Due to the self-similarity property of 
this distribution, if we draw $X$ from a Levy distribution with known 
parameters $X \sim \textrm{Levy}(\mu, \lambda)$, we can obtain any other Levy 
distribution by rescaling and shifting this base distribution: $kX + b \sim 
\textrm{Levy}(k\mu + b,kc)$.

Many other distributions hold this property, allowing stochastic back-propagation rules to be determined for distributions such as the Student's t-distribution, Logistic distribution, the class of stable distributions and the class of generalised extreme value distributions (GEV). Examples of co-ordinate transformations $T(\cdot)$ and the resulsting distributions are shown below for variates $X$ drawn from the standard distribution listed in the first column.

\begin{table}[!h]
\centering
\small
\begin{tabular}{p{1.6cm} l l}
\hline
\textbf{Std Distr.} & T$(\cdot)$ & \textbf{Gen. Distr.}\\
\hline \hline
GEV$(\mu, \sigma, 0)$ & $mX \!+\! b$ & GEV$(m\mu \!+\! b, m \sigma, 0)$ \\
Exp$(1)$ & $\mu \!+\! \beta\! \ln(1 \!+\! \exp(-\!X))$ & Logistic$(\mu, \beta)$ \\
Exp$(1)$ & $\lambda X^{\frac{1}{k}}$ & Weibull$(\lambda, k)$ \\
\hline
\end{tabular}
\end{table}

\section{Variance Reduction using Control Variates}
 \label{ap:var_reinforce}
An alternative approach for stochastic gradient computation is commonly 
based on the method of control variates. We analyse the variance properties 
of various estimators in a simple example using univariate function. We then 
show the correspondence of the widely-known REINFORCE algorithm to the 
general control variate framework.

 \subsection{Variance discussion for REINFORCE}

 The REINFORCE estimator is based on
 \begin{align}
 \nabla_{\theta} \E_p [ f(\xi) ] &= \E_p [ ( f(\xi) - b ) \nabla _{\theta} \log 
 p(\xi | \theta)  ], 
 \label{eq:REINFORCE}
 \vspace{-2mm}
 \end{align}
 where $b$ is a baseline typically chosen to reduce the variance of the 
 estimator. 
 
The variance of 
 \eqref{eq:REINFORCE}  scales poorly 
 with the number of random variables \citep{dayan1995}. To see this 
 limitation, consider functions of the form $ f(\xi) = \sum_{i=1}^K f(\xi_i)$, 
 where 
 each individual term and its gradient has a bounded variance, i.e., $ 
 \kappa_l 
 \leq 
 \textrm{Var}[f(\xi_i)] \leq \kappa_u $ and $ \kappa_l \leq 
 \textrm{Var}[\nabla_{\xi_i}f(\xi_i)] \leq \kappa_u $ for some $0\leq 
 \kappa_{l} 
 \leq \kappa_u$ and assume 
 independent or weakly 
 correlated random variables. Given these assumptions the variance of GBP
 \eqref{eq:bonnet} scales as $ \textrm{Var}[ \nabla_{\xi_i} f(\xi) ] \sim O(1) 
 \label{eq:varGBPMuM} $, while the variance for REINFORCE 
 \eqref{eq:REINFORCE} scales as $  \textrm{Var}\left[  \frac{( \xi_i - \mu_i )}{ 
 	\sigma_i^2 } ( f(\xi) - \E[ f(\xi)   ] ) \right] \sim  O(K). 
 \label{eq:varPolGradMuM} $
 \\ \\
 For the variance of GBP above, all terms in $f(\xi)$ that do not 
 depend on $\xi_i$ have zero gradient, whereas for REINFORCE
 the variance involves a summation over all $K$ terms. Even if 
 most of these terms have zero expectation, they still contribute to the 
 variance of 
 the estimator. Thus, the REINFORCE estimator has the undesirable property 
 that its variance scales linearly with the number of independent random 
 variables in the target function, while the variance of GBP is bounded by a 
 constant. 

The assumption of weakly correlated terms is relevant for 
 variational learning in larger generative models where independence assumptions and structure in 
the 
 variational distribution result in free energies that are summations 
 over weakly correlated or independent terms.
 
\subsection{Univariate variance analysis}
\label{ap:Variance}
In analysing the variance properties of many estimators, we discuss the 
general scaling of likelihood ratio approaches in appendix 
\ref{ap:var_reinforce}. As an example to further emphasise the 
high-variance nature of these alternative approaches, we present a short 
analysis in the univariate case.

Consider a random variable $p(\xi) = \N(\xi| \mu, \sigma^2)$ and  a 
simple quadratic function of the form
\begin{align}
f(\xi) &=  c \frac{\xi^2}{2}. 
\end{align}

For this function we immediately obtain the following variances
\begin{align}
Var[ \nabla_{\xi} f(\xi) ] &= c^2 \sigma^2 \label{eq:varGBPMu} \\
Var[ \nabla_{\xi^2} f(\xi)  ] &= 0 \label{eq:varGBPC} \\
Var[ \frac{( \xi - \mu )}{ \sigma } \nabla_{\xi} f(\xi) ] &= 2 c^2 \sigma^2 + 
\mu^2 c^2 \label{eq:varGBPSigma} \\
Var[  \frac{( \xi - \mu )}{ \sigma^2 } ( f(\xi) - \E[ f(\xi)   ] ) ] &= 2 c^2 
\mu^2 + \frac{5}{2} c^2 \sigma^2 \label{eq:varPolGradMu}
\end{align}
 
 Equations \eqref{eq:varGBPMu}, \eqref{eq:varGBPC} and 
 \eqref{eq:varGBPSigma} correspond to the variance of the estimators based 
 on \eqref{eq:bonnet}, \eqref{eq:price}, \eqref{eq:gradR} respectively 
 whereas equation \eqref{eq:varPolGradMu} corresponds to the variance of 
 the REINFORCE algorithm for the gradient with respect to $\mu$.
 
 From these relations we see that, for any parameter configuration, the 
 variance of the REINFORCE estimator is strictly larger than the variance of 
 the estimator based on \eqref{eq:bonnet}. Additionally, the ratio between 
 the variances of the former and later estimators is lower-bounded by 
 $5/2$. We can also see that the variance of the estimator based on equation
 \eqref{eq:price} is zero for this specific function whereas the variance of 
 the estimator based on equation \eqref{eq:gradR} is not.
 

\section{Estimating the Marginal Likelihood}
\label{sec:app_margLik}
We compute the marginal likelihood by importance sampling by generating $S$ samples from the recognition model and using the following estimator:
\begin{align}
p(\vis) & \approx \frac{1}{S} \sum_{s = 1}^S \frac{p(\vis | 
\vh(\vxi^{(s)})) p(\vxi^{(s)})}{q(\vxi^{s}| \vis )} ; \quad \vxi^{(s)}\sim q(\vxi 
| \vis) \label{eq:FEresults}
\end{align}
 
\section{Missing Data Imputation}
\label{ap:ImageCompletion}
Image completion can be approximatively achieved by a simple iterative 
procedure which consists of (i) initializing the non-observed pixels with 
random values; (ii) sampling from the recognition distribution given the 
resulting image; (iii) reconstruct the image given the sample from the 
recognition model; (iv) iterate the procedure.

We denote the observed and missing entries in an observation as $\vis_o, 
\vis_m$, respectively. The observed $\vis_o$ is fixed throughout, therefore 
all the computations in this section will be conditioned on $\vis_o$.
The  imputation procedure can be written formally as 
a Markov chain on the space of missing entries $\vis_m$ with transition 
kernel $T^q( \vis'_m | \vis_m, \vis_o) $ 
given by
\begin{align}
T^q( \vis'_m | \vis_m, \vis_o ) &= \iint p( \vis'_m, \vis'_o | \xi ) q( \xi| \vis )  d \vis'_o d \xi,
\label{eq:KernelQ}
\end{align}
where $\vis = ( \vis_m, \vis_o )$.

Provided that the recognition model $q( \xi| \vis )$ constitutes a good 
approximation of the true posterior $p( \xi| \vis )$, \eqref{eq:KernelQ} can 
be seen as an approximation of the kernel
\begin{align}
T( \vis'_m | \vis_m, \vis_o ) &= \iint p(\vis'_m, \vis'_o | \xi ) p( \xi| \vis)  d \vis'_o d \xi.
\label{eq:KernelP}
\end{align}
The kernel \eqref{eq:KernelP} has two important properties: (i) it has as its
eigen-distribution the marginal $p(\vis_m| \vis_o)$; (ii) $T(\vis'_m 
|
\vis_m, \vis_o) > 0 ~\forall \vis_o,\vis_m,\vis'_m$. The property (i) can be derived by applying the 
kernel \eqref{eq:KernelP} to the marginal $p(\vis_m| \vis_o)$ and noting 
that it is a fixed point. Property (ii) is an immediate consequence of the 
smoothness of the model.

We apply the fundamental theorem for Markov chains \citep[pp. 38]{neal93} and conclude that given the above
properties, a Markov chain generated by 
\eqref{eq:KernelP} is guaranteed to generate samples from the correct marginal 
$p(\vis_m| \vis_o)$.

In practice, the stationary distribution of the completed pixels will not be 
exactly the marginal $p(\vis_m| \vis_o)$, since we use the 
approximated kernel \eqref{eq:KernelQ}. 
Even in this setting we can provide a bound on the $L_1$ norm of the 
difference between the resulting 
stationary marginal and the target marginal $p(\vis_m| \vis_o)$ 

\begin{proposition}[$L_1$ bound on marginal error ]
If the recognition model $q ( \xi | \vis)$ is such that for all $\xi$
\begin{align}
\exists \varepsilon >0 \textrm{ s.t. }  \int  \left| \frac{q(\xi | \vis) p(\vis)}{p(\xi)} - p ( \vis | \xi ) \right|  d \vis \leq \varepsilon  \label{eq:CondBound}
\end{align}
then the marginal $p(\vis_m|\vis_o)$ is a weak fixed point of the kernel 
\eqref{eq:KernelQ} in the following sense:
\begin{multline}
\int  \Bigg| \int  \big( T^q( \vis'_m | \vis_m, \vis_o )  -  \\
T( \vis'_m | \vis_m, \vis_o )\big)\ p(\vis_m|\vis_o) d \vis_m \Bigg| d \vis'_m < \varepsilon. 
\label{eq:FixBound}
\end{multline}
\end{proposition}

\vspace{0.5cm}
\begin{proof}
\begin{align} 
& \int \!  \left| \! \int \!\left[ T^q( \vis'_m | \vis_m, \vis_o )  \!-\! 
T( \vis'_m | \vis_m, \vis_o )\right] p(\vis_m|\vis_o)  d \vis_m \right | d \vis'_m \nonumber\\
 = & \int  | \iint   p ( \vis'_m, \vis'_o | \xi) p ( \vis_m, \vis_o) [ q ( \xi |
\vis_m, \vis_o)  \nonumber\\
- &   p ( \xi | \vis_m, \vis_o)]d \vis_m d \xi | d \vis'_m 
\nonumber \\
 = & \int  \left | \int   p ( \vis' | \xi) p ( \vis) [ q ( \xi |
\vis) - p ( \xi | \vis)] \frac{p(\vis)}{p(\xi)} \frac{p(\xi)}{p(\vis)} d \vis d \xi 
\right| d \vis'
\nonumber \\
 = & \int  \left | \int  p ( \vis' | \xi) p ( \xi) [ q ( \xi |
\vis) \frac{p(\vis)}{p(\xi)} - p ( \vis | \xi)] d \vis d \xi \right| d \vis' 
\nonumber \\
\leq & \int  \int  p ( \vis' | \xi) p ( \xi) \int   \left | q ( \xi |
\vis) \frac{p(\vis)}{p(\xi)} - p ( \vis | \xi)  \right| d \vis d \xi d \vis' 
\nonumber \\
\leq & \varepsilon, \nonumber
\end{align}
where we apply the condition \eqref{eq:CondBound} to obtain the last 
statement. 
\end{proof}

That is, if the recognition model is sufficiently close to the true posterior 
to guarantee that \eqref{eq:CondBound} holds for some acceptable error 
$\varepsilon$ than \eqref{eq:FixBound} 
guarantees that the fixed-point of the Markov chain induced by the kernel
\eqref{eq:KernelQ} is no further than $\varepsilon$ from the true 
marginal with respect to the 
$L_1$ norm.

%% file: appendix_VB.tex
\section{Variational Bayes for Deep Directed Models}
In the main test we focussed on the variational problem of specifying an posterior on the latent variables only. It is natural to consider the variational Bayes problem in which we specify an approximate posterior for both the latent variables and model parameters.

Following the same construction and considering an Gaussian approximate distribution on the model parameters $\vtheta^g$, the free energy becomes:
\begin{align}
\FE(\vV)
&= -\sum_n
 \overbrace{\E_{q}{ \left[ \log p(\vis_n| \hid(\vxi_n) )   \right]  
}}^{\text{reconstruction error}} \nonumber 
\\
&+ \underbrace{\frac{1}{2} \sum_{n,l} \left[ \Vert \vmu_{n,l} \Vert ^2 + 
\Tr \vC _{n,l}  - \log \vert \vC_{n,l} \vert - 1 \right] }_{\text{latent 
regularization 
term}}\nonumber \\ 
&+\underbrace{\frac{1}{2}\sum_j \left[ \frac{m_j^2}{\kappa} + 
\frac{\tau_j}{\kappa} + \log \kappa - 
\log \tau_j - 1 \right]}_{\text{parameter regularization term}}, 
\label{eq:FE_VB}
\end{align}
which now includes an additional term for the cost of using parameters and their regularisation. We must now compute the additional set of gradients with respect to the parameter's mean $m_j$ 
and variance $\tau_j$ are:

\begin{align}
\nabla_{m_j} \FE(\vis) 
&= -\expect{q}{\nabla_{\theta_j^g} \log p(\vis | \vh(\vxi))} + m_j\\
\nabla_{\tau_j} \FE(\vis) 
&= - \tfrac{1}{2}\E_{q} \left[ \frac{\theta_j - m_j}{\tau_j} 
\nabla_{\theta_j^g} \log p(\vis| \hid(\xi) ) 
\right] \nonumber\\
&+ \frac{1}{2 \kappa} - \frac{1}{2 \tau_j} 
\label{eq:dFEdTau}
\end{align}

%% file: appendix_results.tex
\section{Additional Simulation Details}
We use training data of various types including binary and real-valued 
data sets. In all cases, we train using mini-batches, which requires the 
introduction of scaling terms in the free energy objective function 
\eqref{eq:FEsimp} in order to maintain the correct scale between the 
prior over the parameters and the remaining terms  
\citep{conf/icml/AhnBW12, 
ICML2011Welling_398}. We make use of the objective: 
\begin{align}
& \overline{\FE(\vV)} = -\lambda \sum_n
\E_{q} \left[ \log p(\vis_n| \hid(\vxi_n) ) \right]  
 + \tfrac{1}{2 \kappa} \Vert \vtheta^g\Vert ^2 \nonumber \\
&+  \frac{\lambda}{2} \sum_{n,l} \left[ \Vert \vmu_{n,l} \Vert ^2 + 
\Tr(\vC _{n,l})  - \log \vert \vC_{n,l} \vert - 1 \right], 
\label{eq:FEsimpMiniBatch}
\end{align}
where $n$ is an index over observations in the mini-batch and $\lambda$ is 
equal to the ratio of the data-set and the mini-batch size. At each iteration, a 
random mini-batch of size 200 observations is chosen.

All parameters of the model were initialized using samples from a Gaussian 
distribution with mean zero and variance $1\times 10^6$; the prior 
variance of the parameters was $\kappa = 1\times 10^6$. We compute the 
marginal likelihood on the test data by importance sampling using samples 
from the recognition model; we describe our estimator in appendix 
\ref{sec:app_margLik}. 
%
%
%
%